\documentclass[10pt]{article} 
\usepackage[preprint]{tmlr}
\usepackage{xcolor}
\newcommand{\rev}[1]{{\color{black}#1}}

\usepackage{hyperref}
\usepackage{url}
\usepackage{booktabs}
\usepackage{graphicx}

\usepackage{amsmath,amssymb,amsfonts}
\usepackage{amsthm}
\usepackage[size=small]{caption}

\theoremstyle{definition}
\newtheorem{assumption}{Assumption}

\newcommand{\indep}{\perp \!\!\! \perp}
\newcommand{\muRisk}{\mu\textsf{-risk}}
\newcommand{\tauRisk}{\tau\textsf{-risk}}

\DeclareMathOperator*{\argmin}{argmin}

\title{The Challenges of Hyperparameter Tuning for Accurate Causal Effect Estimation}

\author{\name Damian Machlanski \email d.machlanski@ed.ac.uk \\
      \addr School of Engineering, CHAI Hub\\
      University of Edinburgh
      \AND
      \name Spyridon Samothrakis \email ssamot@essex.ac.uk \\
      \addr Institute for Analytics and Data Science\\
      University of Essex
      \AND
      \name Paul Clarke \email pclarke@essex.ac.uk \\
      \addr Institute for Social and Economic Research \\
      University of Essex}

\def\month{MM}  
\def\year{YYYY} 
\def\openreview{\url{https://openreview.net/forum?id=XXXX}} 

\begin{document}

\maketitle

\begin{abstract}

ML is playing an increasingly crucial role in estimating causal effects of treatments on outcomes from observational data. Many ML methods (`causal estimators') have been proposed for this task. All of these methods, as with any ML approach, require extensive hyperparameter tuning. For non-causal predictive tasks, there is a consensus on the choice of tuning metrics (e.g. mean squared error), making it simple to compare models. However, for causal inference tasks, such a consensus is yet to be reached, making any comparison of causal models difficult. On top of that, there is no ideal metric on which to tune causal estimators, so one must rely on proxies. Furthermore, the fact that model selection in causal inference involves multiple components (causal estimator, ML regressor, hyperparameters, metric), complicates the issue even further. In order to evaluate the importance of each component, we perform an extensive empirical study on their combination. Our experimental setup involves many commonly used causal estimators, regressors (`base learners' henceforth) and metrics applied to four well-known causal inference benchmark datasets. Our results show that hyperparameter tuning increased the probability of reaching state-of-the-art performance in average ($65\% {\rightarrow} 81\%$) and individualised ($50\% {\rightarrow} 57\%$) effect estimation with only commonly used estimators. We also show that the performance of standard metrics can be inconsistent across different scenarios. Our findings highlight the need for further research to establish whether metrics uniformly capable of state-of-the-art performance in causal model evaluation can be found.

\end{abstract}

\section{Introduction}
Modern Machine Learning (ML) methods are remarkably flexible and hence perfectly suited to causal effect estimation, either as sub-components of an overall causal inference procedure (e.g. \citet{chernozhukovDoubleDebiasedMachine2018,kunzelMetalearnersEstimatingHeterogeneous2019}) or as an `algorithmic substrate' upon which further developments take place (e.g. \citet{shalitEstimatingIndividualTreatment2017,atheyGeneralizedRandomForests2019a}). However, the hyperparameters of ML learners must be highly tuned to the dataset at hand to avoid large estimation errors \citep{bergstraAlgorithmsHyperParameterOptimization2011}. In practice, hyperparameter tuning heavily relies on accurate model evaluation, but this task is more challenging for causal than non-causal ML for the fundamental reason that causal parameters depend on unobservable counterfactuals.

Specifically, a key practice in ML is to follow a model selection procedure that reliably identifies the best solutions across a rich space of candidate models and hyperparameters, each of which is evaluated using performance-validation metrics and cross-validation. This task is especially difficult in causal effect estimation because causal parameters are functions of the difference in `potential outcomes’ $\mathcal{Y}_1 - \mathcal{Y}_0$, where $\mathcal{Y}_1$ represents individuals’ outcomes under the treatment and $\mathcal{Y}_0$ those outcomes for the same individuals had they instead not received the treatment:\ hence, one potential outcome is always `counterfactual' and unobservable \citep{robinsROLEMODELSELECTION1986}.

For model evaluation, this means it is only possible to measure prediction error based on the observed outcomes $Y=(1-T)\mathcal{Y}_0 + T\mathcal{Y}_1$, where $T=1$ for individuals who are treated and $T=0$ for individuals who are not. Prediction error is then assessed using e.g. the \textit{observable} Mean Squared Error (oMSE) based on $Y$. Ideally, we would measure prediction error directly on $\mathcal{Y}_1 - \mathcal{Y}_0$, but the corresponding \textit{potential} Mean Squared Error (pMSE) is inaccessible. Were the observed data balanced between treated and untreated units across the entire support of our pre-treatment covariates $X$, as in the case for data from Randomised Controlled Trials (RCT), we could expect oMSE to be an accurate proxy for pMSE. Or, to put it differently, this idealised setup would make it reasonable to assume that the models with the best observed data fit (lowest oMSE) also have the best potential data fit (lowest pMSE). We demonstrate some of the issues arising from such an assumption in Section \ref{sec:problem}.

Observational data, however, does not come from RCTs because treatment selection is non-random and assumed to depend on X. This leads to a domain adaptation, or covariate-shift, problem, as the distribution of $X$ among treated and untreated subjects can be very different \citep{machlanskiUndersmoothingCausalEstimators2022}. As a result, oMSE can be a poor approximation of pMSE, as illustrated in Section \ref{sec:problem}. Because hyperparameter tuning is so crucial for ML methods, bad oMSE can lead to badly tuned ML models and, ultimately, to causal estimators with poor pMSE \citep{qianPerformanceGuaranteesIndividualized2011,doutreligne2025select}.

In this study, we rigorously investigate the influence of hyperparameter tuning on causal estimator performance. In order to achieve this, we look at a) the impact of the quality of hyperparameters on causal estimation performance, and b) effectiveness of observable metrics (oMSE) at approximating ideal but inaccessible target metrics (pMSE) in the tuning task. The role of hyperparameters in causal estimation is understudied. Past research investigated the impact of broader model selection on causal performance, that is, broader in the sense that both models and hyperparameters are part of a single selection search space, making it impossible to analyse the impact of only hyperparameters on performance (e.g. \citet{schulerComparisonMethodsModel2018,mahajanEmpiricalAnalysisModel2022}). At the same time, many papers proposing new causal estimators employ different hyperparameter tuning strategies, which makes it difficult to compare the performance of different estimators in different studies and makes it difficult to assess the influence of tuning on the final performance. We address the issues mentioned above by focusing specifically on hyperparameter selection, and by comparing a large set of existing causal estimators to the best performing estimators on benchmark datasets.

In terms of the effectiveness of observable metrics, there are other studies which have already acknowledged that oMSE can be a poor proxy for pMSE. Some propose new metrics to alleviate the issue \citep{qianPerformanceGuaranteesIndividualized2011,rollingModelSelectionEstimating2014,alaaValidatingCausalInference2019,saitoCounterfactualCrossValidationStable2020a,nieQuasioracleEstimationHeterogeneous2021}, whereas others systematically review currently available oMSE metrics \citep{schulerComparisonMethodsModel2018,mahajanEmpiricalAnalysisModel2022}. However, none of these studies explores the potential performance achievable using pMSE metrics, which is crucial to understanding the impact on practice, because counterfactuals are by definition unobserved. We overcome this problem by leveraging causal inference benchmark datasets, namely, IHDP, Jobs, Twins and News.

These benchmark datasets are designed specifically to test the performance of causal estimators and hence include the ground truth in the form of either counterfactuals or causal effects, making metrics like pMSE available to us. However, note that the training datasets used herein comprise only observational data (i.e., no counterfactuals or causal effects) such that the ground truth and pMSE metrics are used only for performance assessment purposes and not for model selection/tuning.

Our interest in model selection and evaluation metrics combined with the empirical nature of this study create a similar context to that found in two of the papers mentioned above. \citet{schulerComparisonMethodsModel2018} compare the effectiveness of multiple observable metrics in model selection using simulated data. \citet{mahajanEmpiricalAnalysisModel2022} perform a similar analysis but attempt to make their datasets more realistic through generative modelling \citep{nealRealCauseRealisticCausal2021a}, a concept previously proposed by \citet{schulerSynthValidationSelectingBest2017}. Both are useful for identifying the best model selection metrics and for observing how model selection affects estimation performance. Our work differs in two important aspects. First, in some of our experiments \rev{we exclusively study the impact of hyperparameters on the performance of individual causal estimators (see a) above) by restricting the model-selection search spaces to involve only hyperparameters.} Second, we include ideal pMSE metrics in model selection to obtain potential performances. This allows us to better understand the magnitude of the consequences of oMSE inaccurately approximating pMSE (important for hyperparameter tuning), and explore potential performances of causal estimators given the right hyperparameters (see b) above).

Before introducing the study design and presenting the results, we summarize the key empirical findings of our study below:
\begin{itemize}
    \item The right hyperparameters, \rev{selected} with ideal/potential metrics, \rev{\textbf{significantly increase} the chances of achieving state-of-the-art (SotA) performance \textbf{across most} types} of causal estimators and base learners, provided that the \rev{chosen} ML learners are flexible enough \rev{for} the problem at hand.
    \textit{Conclusion: choosing correct hyperparameters is more important than selecting causal estimators and base learners.}
    \item For most common causal estimators, there exists a set of suitable hyperparameters that \textbf{reach SotA} performance based on observed metrics for other estimators and learners.
    \textit{Conclusion: hyperparameters can be used to achieve arbitrary estimation performance, questioning recent benchmarks.}
    \item The quality of hyperparameters selected with different commonly used \textit{observable} metrics \textbf{varies substantially}.
    \textit{Conclusion: the choice of (observable) metric used in hyperparameter tuning is highly influential on estimation performance.}
    \item \rev{Some} commonly used \textit{observable} metrics \rev{may be \textbf{highly variable} and lead to \textbf{inconsistent results}.}
    \textit{Conclusion: \rev{more robust} observable metrics are needed \rev{to achieve} excellent \rev{estimation} performances with consistency.}
\end{itemize}

The rest of the paper is structured as follows. Section \ref{sec:prelim} gives a brief background overview on the topics of causal effect estimation and model selection, followed by Section \ref{sec:problem} discussing the challenges of model evaluation in the causal effect estimation setting. Proposed methodology is described in Section \ref{sec:meth}. Obtained results are presented and discussed in Section \ref{sec:results}, with Section \ref{sec:conclusion} concluding the paper.

\section{Preliminaries}\label{sec:prelim}
A brief overview of the two topics fundamental to this work is given in the following subsections. These are the problems of causal effect estimation and causal model selection. For a thorough discussion on the topics, refer to \citep{yaoSurveyCausalInference2020, guoSurveyLearningCausality2020} and \citep{rollingModelSelectionEstimating2014, schulerComparisonMethodsModel2018} respectively, as well as \citep{pearl2009causality} for a general causal analysis background.

\subsection{Causal Effect Estimation}
Using Rubin's potential outcome framework \citep{rubinEstimatingCausalEffects1974}, the problem of estimating causal (treatment) effects can be defined as follows. First, the main three variables of interest are background covariates $X$, treatment assignment $T$, and outcome $\mathcal{Y}$. Second, a potential outcome $\mathcal{Y}_t^{(i)}$ is the observed outcome when individual $i$ receives treatment $t=0,1$. In addition, it is customary to make a set of assumptions about the underlying data generating process, which we also employ in this study.

\begin{assumption}[Stable Unit Treatment Value Assumption (SUTVA)]
The potential outcomes of any unit \rev{depend only on} the treatment \rev{administered to that unit but are unaffected by the treatment administered to any other unit.} Furthermore, there are no different levels or forms of the same treatment.
\end{assumption}

\begin{assumption}[Ignorability]
Given background covariates $X$, potential outcomes $\mathcal{Y}$ are independent from observed treatment $T$. That is, $\mathcal{Y}_1, \mathcal{Y}_0 \indep T \mid X$.
\end{assumption}

\begin{assumption}[Positivity]
Treatment assignment $T$ is not deterministic for all individuals $X$. That is, $P(T=t \mid X=x)>0$ for all $t$ and $x$.
\end{assumption}

In practice, they translate to a) no inter-unit interaction and no hidden treatment variations (SUTVA), b) for individuals with the same $X$, their treatment assignment $T$ can be perceived as random because there are no unmeasured confounders (ignorability) thus often referred to as \textit{unconfoundedness}, and c) potential outcomes for all treatments and units (and their combinations) are required (positivity). In addition, ignorability and positivity together form \textit{strong ignorability}. Given the definitions and assumptions, the Individual Treatment Effect (ITE) can be written as:
\begin{equation}\label{eq:ite}
    ITE^{(i)} = \mathcal{Y}_1^{(i)} - \mathcal{Y}_0^{(i)}
\end{equation}

Such quantities are, however, not identifiable by observed data (missing counterfactuals), unlike Conditional Average Treatment Effect (CATE) and Average Treatment Effect (ATE) parameters, defined as:
\begin{equation}\label{eq:cate}
    CATE(x) = \tau(x) = \mathbb{E}\left [ \mathcal{Y}_1 \mid X=x \right ] - \mathbb{E}\left [ \mathcal{Y}_0 \mid X=x \right ]
\end{equation}
\begin{equation}
    ATE = \mathbb{E}\left [ \tau(X) \right ] = \mathbb{E}\left [ \mathcal{Y}_1 - \mathcal{Y}_0 \right ]
\end{equation}
where $\mathbb{E}[.]$ denotes mathematical expectation. Thus, ATE is a population-level average ITE, whereas CATE is average ITE for a subpopulation $X=x$. Moreover, CATE is more meaningful than, and thus preferred, to ATE in scenarios where there is a considerable degree of ITE heterogeneity and this heterogeneity varies between subpopulations.

\subsection{CATE Estimators}
Many estimators are available with which to estimate ATE under the above assumptions. Arguably, the most intuitive approach is regression adjustment involving a single regression learner $\mu(t,x) = \mathbb{E}[\mathcal{Y} \mid T=t,X=x]$. Recognising the need for different modelling complexity per treatment arm, it is natural to extend the approach to two separate learners $\mu_t(x) = \mathbb{E}[\mathcal{Y} \mid T=t,X=x]$ for $t=0$ and $t=1$. Single- and two-learner regression adjustment approaches have been formalised as S- and T-Learners respectively (e.g.\ \citet{kunzelMetalearnersEstimatingHeterogeneous2019}). Another method of adjustment is to use the \textit{propensity score} $e(x) = P(T=1 \mid X=x)$, that is, the probability of a unit receiving the treatment. The most straightforward way to incorporate response propensities is through Inverse Propensity Score Weights (IPSW) to counteract data imbalances caused by non-random selection bias and covariate shifts \citep{rosenbaumCentralRolePropensity1983}. In practice, both $\mu(x)$ and $e(x)$ can be subject to model misspecification. This led to the development of the Doubly Robust estimators \citep{robinsEstimationRegressionCoefficients1994} which allow consistent estimation of target parameters (the property where the estimator converges to the truth as the sample size increases to infinity) even if one (but not both) of nuisance models $\mu(x)$ and $e(x)$ is misspecified. The Targeted Maximum Likelihood Estimation (TMLE) approach builds on this idea to incorporate ML for the nuisance models through ‘super-learning’ and uses influence functions derived from the theory of statistical functionals to minimise finite-sample bias \citep{laanTargetedMaximumLikelihood2006}.

The literature on CATE estimation is more recent. Doubly Robust and TMLE estimators can be extended for CATE but require users to specify a family of parametric models. Double Machine Learning (DML) \citep{chernozhukovDoubleDebiasedMachine2018} introduces a general framework for parametric CATEs, implementation of which can assume linear effects \citep{econml} or be applied to manually specified non-linear cases. DML additionally proposes the use of Neyman-orthogonal estimating equations for CATEs. These have the property of limiting the impact of bias from estimation of the nuisance parameters ($\mu(x)$ and $e(x)$) which can be minimized when combined with cross-fitting. Hence, DML proposes two-stage estimation: the first stage involves estimating nuisance functions $\mu(x)$ and $e(x)$; then, the residuals $\mathcal{Y} - \mu(x)$ and $T - e(x)$ are used to minimise the residual on the CATE square loss. More recent work generalizes this two-stage approach to the R-Learner with which CATE can be modelled non-parametrically \citep{nieQuasioracleEstimationHeterogeneous2021}. All of these approaches fall into the class of Orthogonal Learning methods \citep{fosterOrthogonalStatisticalLearning2020}. A different approach but also involving multi-stage modelling is X-Learner \citep{kunzelMetalearnersEstimatingHeterogeneous2019}. This is shown to be a special case of the R-Learner for problems where one treatment arm (usually the untreated/control) dominates the other in terms of sample size: it behaves as T-Learner in the first stage, models \textit{imputed effects} $\mu_1(X_0) - \mathcal{Y}_0$ and $\mathcal{Y}_1 - \mu_0(X_1)$ in stage two, with the third stage being final CATE estimation weighted by propensity $e(x)$ \citet{nieQuasioracleEstimationHeterogeneous2021}.

CATE estimators outside of Orthogonal Learning have also been explored. In the realm of tree-like algorithms, there is Causal Forest \citep{atheyGeneralizedRandomForests2019a} that generalises Decision Trees and Random Forests to CATE estimation. Unlike standard trees which splits are based on, for instance, MSE or entropy, causal trees are based on CATE heterogeneity, resulting in the tree leaves containing both treated and control samples but assigned to the same heterogeneous group. Within-group difference then enables CATE estimation. Using ensembles of trees, one gets Causal Forests, which offer a unique approach to CATE estimation, though their non-standard design makes hyperparameter tuning more challenging as it renders standard performance metrics unusable. Trees have also been used as a data augmentation tool to address data imbalances via Generative Trees for improved CATE estimation \citep{machlanskiUndersmoothingCausalEstimators2022}.

Neural Networks (NN) have also been used in CATE estimation, many of which established new SotA performance levels. Some of the more important ideas involve penalising imbalanced representations using standard feedforward networks \citep{johanssonLearningRepresentationsCounterfactual2016}, further extended to two-headed structures (one head per treatment group) and more sophisticated distribution discrepancy metrics \citep{shalitEstimatingIndividualTreatment2017}, as well as solutions that encourage the preservation of local similarities on top of balanced representations \citep{yaoRepresentationLearningTreatment2018}. In particular, the two-headed structure without penalisation, termed TARNet \citep{shalitEstimatingIndividualTreatment2017}, inspired many other approaches, combining it with TMLE \citep{shiAdaptingNeuralNetworks2019b} or ensemble learning \citep{samothrakisGrokkinglikeEffectsCounterfactual2022}. Other methods incorporate generative NNs in the form of Variational Autoencoders \citep{louizosCausalEffectInference2017a} and Generative Adversarial Networks \citep{yoonGANITEEstimationIndividualized2018}.

\subsection{Model Selection}\label{sec:ms}
No single model is universally better than others in all situations (\textit{no free lunch} theorem). Thus, a model most suitable to the task at hand must be selected every time a new task is encountered, either manually by employing prior knowledge about the problem, or in a data-driven manner, the latter being the focus of this work. More formally, the task is to select a model $\mathcal{M}_{m^*}$ from a collection of candidate models $\mathcal{M} = \{\mathcal{M}_1, ..., \mathcal{M}_M\}$ that minimises incurred loss $\mathcal{L}$ on validation data $\mathcal{D}^{val}$ after being trained on training data $\mathcal{D}^{tr}$. To simplify the notation, let us also introduce a model $\mathcal{M}_m$ trained on training data $\mathcal{D}^{tr}$ as a predictive model $\mathcal{P}_m$. Putting both equations together, we have

\begin{equation}\label{eq:model}
\mathcal{P}_m = \mathcal{P}(\mathcal{M}_m, \mathcal{D}^{tr})
\end{equation}
\begin{equation}\label{eq:ms}
m^* = \argmin_{m=1,...,M}{\mathcal{L}(\mathcal{P}_m, \mathcal{D}^{val})}
\end{equation}

To reduce the clutter, let us refer to winning models $\mathcal{M}_{m^*}$ and $\mathcal{P}_{m^*}$ as simply $\mathcal{M}^*$ and $\mathcal{P}^*$ respectively. The goal could also be adjusted to maximise a scoring criteria, instead of minimising a loss function. Note, candidate models $\mathcal{M}$ may include different CATE estimators (e.g. X-Learner, Doubly Robust), base learners (e.g. Linear Regression, Decision Tree) and hyperparameters (e.g. regularisation strength, hidden layers). Therefore $\mathcal{M}$ can be defined as a tuple $\{\mathcal{C},(\mathcal{B},\mathcal{H})\}$ that consists of causal estimators $\mathcal{C}$ and a pair of base learners and hyperparameter values $(\mathcal{B, H})$, with $c \in \Omega_{\mathcal{C}}$ and $(b,h) \in \Omega_{\mathcal{B,H}}$. $\mathcal{B}$ and $\mathcal{H}$ must be a pair as not all base learner-hyperparameter combinations are meaningful. Thus, it is clear that changing any of the three will result in a new candidate model. An \textit{S-Learner} with \textit{Linear Regression} as its base learner and a hyperparameter $L1=0.1$ could be one possible example of a candidate model, say $\mathcal{M}_1$. Then, using the same CATE estimator and base learner but changing the hyperparameter to $L1=0.5$ now results in a different model $\mathcal{M}_2$. This is because, even though both models have the same $\mathcal{C}_c$ and $\mathcal{B}_b$, they differ in $\mathcal{H}_h$, hence they are different candidate models ($\mathcal{M}_1 \neq \mathcal{M}_2$). Rewriting the problem to include all three search spaces explicitly results in
\begin{equation}\label{eq:search}
m^* = \{c^*,(b^*,h^*)\} = \argmin_{
                        \substack{
                        c=1,...,C \\
                        b=1,...,B \\
                        h=1,...,H
                        }
                        }
                        {\mathcal{L}(\{\mathcal{C}_c, (\mathcal{B}_b, \mathcal{H}_h)\}, \mathcal{D}^{val})}
\end{equation}

Note that many causal estimators allow for multiple base learners to be defined, so a single $\mathcal{B}_b$ may contain a tuple of multiple learners. Similarly, many base learners have multiple tunable hyperparameters, hence it is acceptable for a single $\mathcal{H}_h$ to be defined as a set of multiple hyperparameters. The loss $\mathcal{L}$ is commonly obtained on a single validation set or through cross-validation (CV), where the final loss is an average of losses obtained on each validation fold. For this reason, $\mathcal{D}^{tr} \cap \mathcal{D}^{val}$ is not necessarily $\varnothing$ (i.e. empty) in case CV is used.

\subsection{Observable and Potential Metrics}
We further categorise loss $\mathcal{L}$ into two classes depending on the type of data it has access to. The \textit{observed} loss $\mathcal{L}_{obs}$ has access to observational data $\mathcal{D}_{obs} = \{ X,T,\mathcal{Y}_t \}$, where $X$ denotes background covariates, $T$ treatment status, and $\mathcal{Y}_t$ observed outcomes.
\begin{equation}\label{eq:l_obs}
    \mathcal{L}_{obs}^m = \mathcal{L}(\mathcal{P}_m, \mathcal{D}_{obs})
\end{equation}

Whereas the \textit{potential} loss $\mathcal{L}_{pot}$ uses the same data, but has access to all potential outcomes $\mathcal{Y}$, together forming potential data $\mathcal{D}_{pot} = \{ X,T,\mathcal{Y}_0,\mathcal{Y}_1 \}$.
\begin{equation}\label{eq:l_pot}
    \mathcal{L}_{pot}^m = \mathcal{L}(\mathcal{P}_m, \mathcal{D}_{pot})
\end{equation}

\subsection{Evaluation of Model Selection}\label{sec:eval_ms}
In addition to model evaluation for model selection purposes, assessing effectiveness of an evaluation metric at the task could also be desirable, especially when comparing multiple metrics. In order to avoid overfitting, evaluation of model selection metrics should be performed on a separate set of data than model selection itself. For this purpose, let us define validation and test data, $\mathcal{D}^{val}$ and $\mathcal{D}^{te}$ respectively. Note that $(\mathcal{D}^{tr} \cup \mathcal{D}^{val}) \cap \mathcal{D}^{te} = \varnothing$ is always true, even in the case of using CV. From now on, we assume model selection is always done on validation data, and the performance assessment of the metric used in model selection is done on a separate test set, unless explicitly stated otherwise.

Following Equations \eqref{eq:model} and \eqref{eq:ms}, we define $\mathcal{P}^*_{obs}$ as the best-performing model evaluated on $\mathcal{D}^{val}_{obs}$ using $\mathcal{L}_{obs}$. Then, we measure the performance of $\mathcal{L}_{obs}$ as the performance of the model $\mathcal{P}^*_{obs}$ but evaluated on $\mathcal{D}^{te}_{pot}$ using $\mathcal{L}_{pot}$.
\begin{equation}\label{eq:ms_val}
    \mathcal{L}_{pot}^* = \mathcal{L}_{pot}(\mathcal{P}^*_{obs}, \mathcal{D}^{te}_{pot})
\end{equation}
Therefore, the ultimate goal becomes to optimise for the potential loss indirectly through model selection performed via the observable loss and data. More generally, when considering multiple observable loss functions $\mathcal{L}_{obs} = \{ \mathcal{L}_1,...,\mathcal{L}_V \}$, and a single potential loss function $\mathcal{L}_{pot}$ of choice, the task can be defined as a nested optimisation problem, wherein the goal is to select an observable loss function $\mathcal{L}_{v^*}$ that, through its model selection, optimises (approximates) the potential loss the best. That is
\begin{equation}\label{eq:sel}
v^* = \argmin_{v=1,...,V}{\mathcal{L}_{pot}(\mathcal{P}^*_v, \mathcal{D}^{te}_{pot})}
\end{equation}

Acknowledging $\mathcal{L}_{pot}$ as the desired ultimate goal, as shown by Equations \eqref{eq:ms_val} and \eqref{eq:sel}, creates the need to incorporate as much information about $\mathcal{L}_{pot}$ into $\mathcal{L}_{obs}$ as possible. In practice, however, $\mathcal{L}_{pot}$ may require access to type of data unavailable to $\mathcal{L}_{obs}$ (e.g. counterfactuals in \textit{PEHE}), making the task extremely challenging. This important realisation of targeting $\mathcal{L}_{pot}$ through available data is foundational to \textit{targeted learning} estimators \citep{laanTargetedMaximumLikelihood2006} and evaluation metrics \citep{alaaValidatingCausalInference2019}.

\subsubsection{Oracle}\label{sec:oracle}
As observable metrics are only imperfect proxies for the potential ones, it is useful to compare their model selection performance to unbiased model selection that is possible via potential metrics. Such unbiased model selection is defined as:
\begin{equation}\label{eq:oracle}
m^{**} = \argmin_{m=1,...,M}{\mathcal{L}_{pot}(\mathcal{P}_m, \mathcal{D}^{val}_{pot})}
\end{equation}

Model $\mathcal{M}_{m^{**}}$, $\mathcal{M}^{**}$ for short, thus is a candidate model that optimises the potential loss $\mathcal{L}_{pot}$ the best among all candidate models $\mathcal{M}$. Note that $\mathcal{M}^{**}$ may differ across different potential metrics $\mathcal{L}_{pot}$. Furthermore, its loss $\mathcal{L}^{**}_{pot}$ with respect to the same potential loss function $\mathcal{L}_{pot}$ is the best possible among defined collection of candidate models.
\begin{equation}\label{eq:oracle_perf}
    Oracle_{weak} = \mathcal{L}^{**}_{pot} = \mathcal{L}_{pot}(\mathcal{P}^{**}_{pot}, \mathcal{D}^{te}_{pot})
\end{equation}

Note that potential data $\mathcal{D}_{pot}$ can contain ground truth information the observable set does not have, such us true causal effects. Thus, in some sense, $\mathcal{M}^{**}$ can be perceived as optimal model choice, at least with respect to the chosen potential metric $\mathcal{L}_{pot}$ and available test set. Due to this relative optimality, we refer to Equation \eqref{eq:oracle} as \textit{Oracle} model selection, and the loss $\mathcal{L}^{**}_{pot}$ as \textit{Oracle} performance. 

An Oracle defined on such terms would arguably be a \textit{weak} one due to it being derived from sample, not population, data. However, if we repeat Equation \eqref{eq:oracle_perf} over $n$ datasets that come from the same distribution, the Oracle will now reflect population data and hence can be considered a \textit{strong} Oracle.
\begin{equation}\label{eq:oracle_str}
    Oracle_{strong} = \mathcal{L}^{**}_{pot} = \frac{1}{n} \sum_i^n \mathcal{L}_{pot}(\mathcal{P}^{**}_{pot}(i), \mathcal{D}^{te}_{pot}(i))
\end{equation}
As our experiments involve calculating strong Oracles, we refer to them as just \textit{Oracles} throughout the rest of the text.

\subsection{Causal Model Selection Methods}
When it comes to model performance evaluation, there are multiple approaches to choose from. Perhaps the most straightforward and most prevalent in supervised learning is MSE between predicted and factual outcomes (see \ref{MSE} below), also referred to in the literature as $\mu\textsf{-risk}$ \citep{schulerComparisonMethodsModel2018} or \textit{factual validation} \citep{alaaValidatingCausalInference2019}. The major shortcoming is the assumption that prediction error on factual outcomes is an accurate proxy for prediction error on CATEs, which, as shown by \citet{rollingModelSelectionEstimating2014} and in Section \ref{sec:problem}, can have severe limitations. That is, it does not target CATE, only $\mu_0(x)$ and $\mu_1(x)$ separately. Nevertheless, its ease of use and popularity in machine learning makes it a common metric of choice, similarly to $R^2$.

In terms of selection methods tailored specifically to CATE estimation, the usual goal is to work around the issue of missing ground truth (see \ref{PEHE} below). Many approaches attempt to synthesise it through another layer of CATE modelling by treating it as a proxy for the final performance, also called \textit{plugin validation} \citep{alaaValidatingCausalInference2019}. More concretely, an additional CATE estimator is trained specifically on validation data, and then provides CATE predictions $\tilde{\tau}(x)$ on the same subset of data. Cross-fitting is often used here to avoid overfitting. Any other candidate CATE estimator that is subject to model evaluation is trained on training data and makes CATE predictions $\hat{\tau}(x)$ on validation data. To evaluate the candidate model, its predictions $\hat{\tau}(x)$ are compared to the synthetic ground truth $\tilde{\tau}(x)$, based on which model selection decisions can be made. The semi-ground truth $\tilde{\tau}(x)$ is essentially \textit{plugged in} instead of true CATEs $\tau(x)$ into, for example, \textit{PEHE} formula (see \ref{PEHE}). Moreover, some solutions involve predicting only counterfactual outcomes and obtaining $\tilde{\tau}(x)$ by reusing given factuals, also referred to as \textit{imputed effects} \citep{kunzelMetalearnersEstimatingHeterogeneous2019}, whereas other approaches do not use factual outcomes and predict $\tilde{\tau}(x)$ directly. Many different models have been used so far as a source of $\tilde{\tau}(x)$. One example is using matching with a distance measure between data points to identify counterfactuals \citep{rollingModelSelectionEstimating2014}, or any other CATE estimator \citep{schulerSynthValidationSelectingBest2017}, including neural networks \citep{saitoCounterfactualCrossValidationStable2020a}, all of which provide synthetic CATEs as a semi-ground truth. Generative models have also been used to generate artificial potential outcomes and subsequently $\tilde{\tau}(x)$ \citep{nealRealCauseRealisticCausal2021a,atheyUsingWassersteinGenerative2020}, with the major drawback being notorious instability of such models. However, the idea of synthesising CATEs shares the same inherent issue with CATE estimation itself, that is, missing counterfactuals. This leaves the problem of model tuning of the CATE synthesiser open. It is also arguably a paradoxical approach, as having a model that accurately synthesises CATEs would solve the main problem in the first place, rendering any further model selection pointless.

A response to possibly biased plugin validation is unplugged validation via influence functions \citep{alaaValidatingCausalInference2019}. Another alternative is \textit{R-Loss} derived from the main formula of the R-Learner \citep{nieQuasioracleEstimationHeterogeneous2021}, further extended to \textit{R-Score} \citep{econml} that penalises constant CATE predictions, similarly to $R^2$ penalising constant outcome predictions in general. Interestingly, it is also possible to rewrite \textit{R-Loss} as a reweighted ideal metric, such as \textit{PEHE} \citep{doutreligne2025select}.

\section{Problem Demonstration}\label{sec:problem}
\begin{figure}[t]
    \centering
    \includegraphics[width=0.99\textwidth]{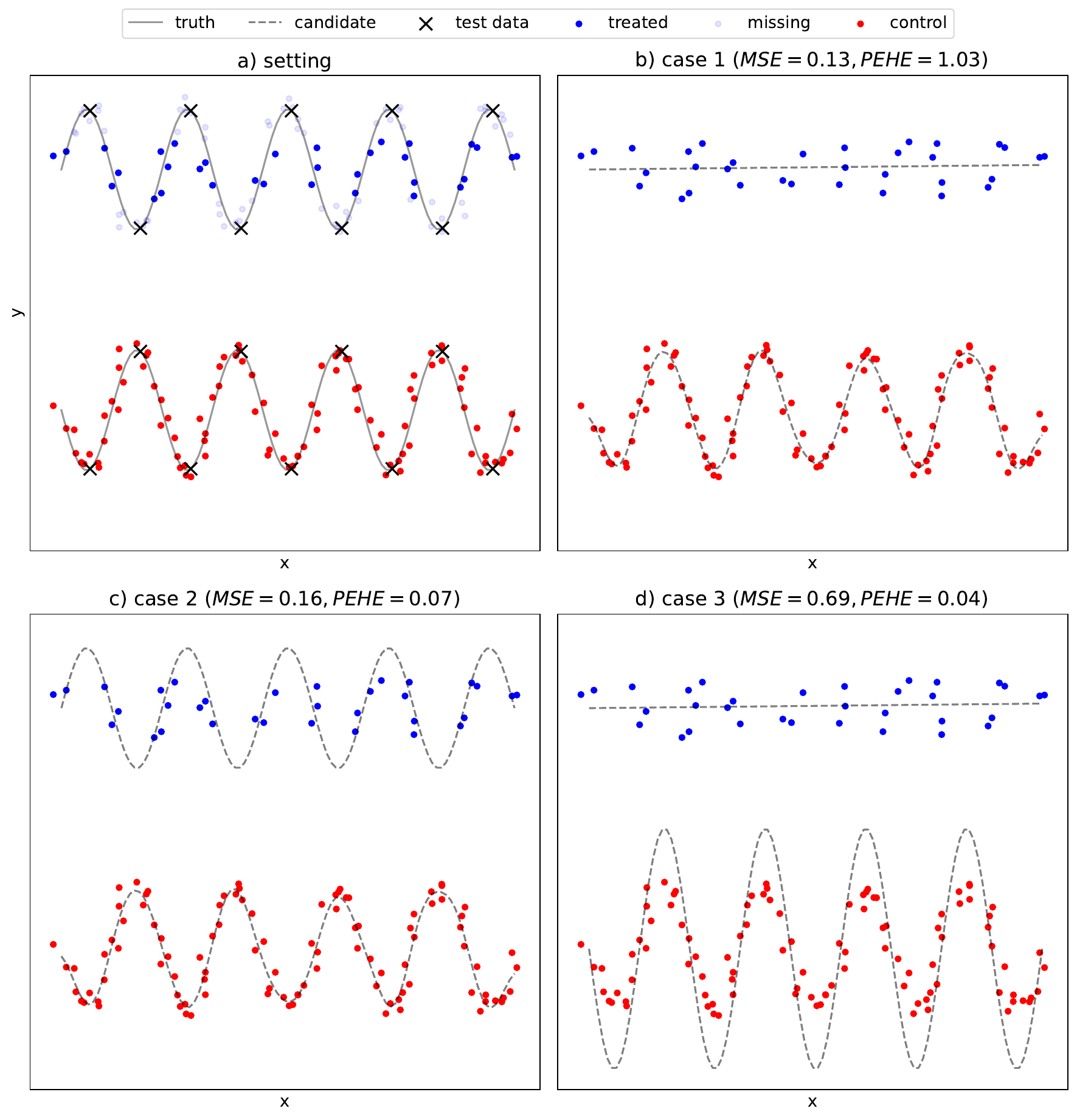}
    \caption{MSE - evaluation metric on observed data (validation set/cross-validation) used for model selection purposes (accessible with real datasets; lower is better). PEHE - evaluation metric on unobserved test data (not accessible with real datasets due to missing counterfactuals; lower is better).}
    \label{fig:example}
\end{figure}

The graphical example in Figure \ref{fig:example} depicts some of the challenges of model selection, or model evaluation in general, in causal effect estimation. It is intentionally extreme to demonstrate the issues clearly and make it easier for the reader to appreciate the difficulty of the task. The scenarios presented here are not all realistic but chosen to demonstrate the potential identification issues that arise because the counterfactual data are unavailable. More concretely, we show how systematically missing data in small samples can affect, with a varying degree, goodness-of-fit and causal metrics and possible further consequences of that.

Our example is a straightforward two-dimensional case, consisting of an input feature $X$, response $Y$ (both continuous), and binary ($0,1$) treatment assignment $T$. The Data Generating Process (DGP) is a non-linear additive noise model where the mean of $Y$ is sinusoidal in $X$ with independent and identically distributed normal errors. Treated cases are often challenging to collect in practice (e.g. high costs, ethical reasons), leading to those units being not fully representative of DGP behind the treated arm. In general, all sorts of imbalances can arise from covariate shift and non-random selection, making the task of modelling the DGP with available data certainly non-trivial. Taken to the extreme, mostly for demonstration purposes, Subfigure a) depicts a possible scenario we can find ourselves in (note faded treated units). Despite the missing data problem, which in practice can exist in both treatment arms, the goal is to model the underlying DGP as close as possible and provide accurate treatment effect predictions not only for observed units, but also the ones the model have not seen in training (see data points marked with `X').
 The error for this is captured via the PEHE metric (\ref{PEHE} below), which is generally inaccessible due to missing counterfactuals in real datasets, but useful in simulations and benchmark datasets designed specifically to assess the performance of CATE estimators where PEHE practically becomes the metric to beat. The crucial difference between the two is that MSE measures average prediction error on observed outcomes $\mathcal{Y}$, whereas PEHE quantifies errors on predicted CATEs (i.e. difference between outcomes $\mathcal{Y}_1 - \mathcal{Y}_0$). That is, given an outcome prediction model $\hat{\mu}_t$:
\begin{equation}\label{MSE}
    MSE = \frac{1}{n}\sum_{i=1}^{n}( \mathcal{Y}_t^{(i)} - (1-t)\hat{\mu_0}(x^{(i)}) - t\hat{\mu_1}(x^{(i)}) )^2
\end{equation}
\begin{equation}\label{PEHE}
    PEHE=\sqrt{\frac{1}{n}\sum_{i=1}^{n}[ (\hat{\mu}_1(x^{(i)})  - \hat{\mu}_0(x^{(i)})) - (\mathcal{Y}_1^{(i)} - \mathcal{Y}_0^{(i)}) ] ^2}
\end{equation}

Note PEHE directly targets $\tau(x) = \mu_1(x)-\mu_0(x)$ and uses both $\mathcal{Y}_0$ and $\mathcal{Y}_1$ (one of them is never observed), whereas MSE targets each of $\mu_0(x)$ and $\mu_1(x)$ individually rather than the difference between them and involves only observed outcomes $\mathcal{Y}_t$. Clearly, PEHE cannot be targeted directly using only the observed data, only MSE; hence, it is tempting to assume that more accurate (in terms of MSE) outcome predictions (better fit) will lead to accurate CATE estimates. However, the disparity between model selection (MSE) and desired performance (PEHE) metrics combined with the aforementioned missing data issue can lead to undesirable solutions, as demonstrated by Subfigures b) - d) and further discussed in the following paragraphs.  Even at this preliminary stage, it becomes relatively apparent that model selection task, which is perceived as a rather solved problem in supervised ML, is a non-trivial challenge in treatment effect estimation setting.

For \textbf{Case 1} (Subfigure b)), the control data are fitted well, accurately capturing the DGP for this part of data. The model in the treated arm failed to match the underlying trend due to severity of data loss. This solution is likely favoured if using standard model selection metrics, such as MSE. Despite the best MSE and winning over the other two candidate solutions, this variant is not desirable due to poor generalisation in the treated arm and overall high CATE prediction errors, as correctly captured by high PEHE.

For \textbf{Case 2} (Subfigure c)), both treatment arm models capture the true DGP very well. Selecting a more complex model that matches the DGP trend in the treated arm despite data loss is especially difficult with conventional performance metrics (see worse MSE than case 1). One possibility for a performance metric to overcome this is to consider the task in its entirety (CATE estimation) instead of fitting each arm separately. Or put differently, the information about the trend in one treatment arm can be useful to modelling choices of the other arm. This way of thinking leads to targeted learning \citep{laanTargetedMaximumLikelihood2006} and in general to more sophisticated causal model selection methods \citep{alaaValidatingCausalInference2019, nieQuasioracleEstimationHeterogeneous2021}. Due to better generalisation, case 2 solution is clearly preferred (better PEHE than case 1), but certainly non-trivial to identify (worse MSE than case 1).

In \textbf{Case 3} (Subfigure d)), we have an example that is much harder to imagine in real life situations, but can certainly happen when the goal is to beat SotA methods with respect to PEHE on causal benchmark datasets. These datasets are often designed specifically to assess the performance of CATE estimators and thus include true CATEs to enable such tests (PEHE is then possible to calculate). While optimising for PEHE directly during the fitting process of CATE estimators would be an unacceptable violation of good practices, it is practically impossible to eliminate the exposure to PEHE completely to zero. For instance, if one was to develop a new CATE estimator, checking PEHE on the test set throughout the development might be an unavoidable step to ensure low enough PEHE is achieved for the new estimator to be considered worth paying attention to by the community and increase overall chances of it being published. Due to such external judgements, the information about the test PEHE may leak into the design of the new CATE estimator. A possible result is an undesirable solution that misses the main DGP trends and thus generalises poorly (the worst MSE of all 3 cases), but because of the information leak about test PEHE, it manages to win the competition in terms of PEHE minimisation (and thus becoming a new SotA CATE estimator on the benchmark). Examining the plots of course makes it easy to discard this particular solution, but for high-dimensional $X$ graphical representations quickly become non-trivial and hard to interpret.

Examining the three example cases together, it can be observed that MSE is a rather poor proxy for PEHE, which is quite counterintuitive as better fit is believed to generally lead to improved CATE estimates. Therefore, simple model selection metrics, such as MSE, may not be sufficient to identify desirable solutions. Focusing too much on benchmark datasets and (indirectly) PEHE can be dangerous as well. Just because a solution achieves a low level of errors on CATEs does not necessarily mean this is the desirable solution. Let us consider the following hypothetical scenario. A new CATE estimator has been developed (case 2). Practitioners perform model selection with MSE and thus choose to proceed with case 1 model due to lower MSE. On the other hand, research community deemed case 2 model not good enough as case 3 estimator achieved better PEHE on benchmark datasets. Both communities rejected a perfectly sound and desirable case 2 solution. How do we overcome this situation? Shall we exploit the benchmarks too?

It is straightforward to pick case 2 solution by examining the plots. With actual datasets, plotting is not an option due to high-dimensionality of input features, leaving performance metrics as the only feedback signal. These are clearly not ideal as relying on any of them exclusively may result in undesirable solutions, as shown in the examples. Perhaps an untapped potential lies in a combination of goodness-of-fit and CATE loss measures, or variations of such. This is motivated by an observation that even though the non-trivial case 2 solution has neither the best MSE, nor the best PEHE, its sum of MSE and PEHE is the lowest. As evidenced by the presented simulation study and analysis, model selection in the causal effect estimation setting is clearly non-trivial and poses many challenges. Performance metrics seem to play an important role, possibly having a much stronger influence on the final performance of CATE estimation than it is commonly assumed. This motivates further study on causal model selection, CATE estimators, and their interplay, but on datasets closer to reality. This is the subject of the remaining content of this paper.

\section{Methodology}\label{sec:meth}
The goal of our experimental setup is to empirically investigate: a) the impact of hyperparameter tuning (and lack thereof) on causal estimation performance, and b) effectiveness of observable metrics (oMSE $=\mathcal{L}_{obs}$) at approximating ideal but inaccessible target metrics (pMSE $=\mathcal{L}_{pot}$) in the task of model selection and tuning. 

We follow the notation used in Sections \ref{sec:ms}-\ref{sec:eval_ms}. Thus, the goal a) is to explore how different hyperparameters across the search space $\mathcal{H}$ affect estimation performance $\mathcal{L}_{pot}$ of estimators $\mathcal{C}_c$ and base learners $\mathcal{B}_b$. Whereas goal b) is about comparing performances $\mathcal{L}_{pot}^*$ achieved via observable metrics $\mathcal{L}_{obs}$ (Equation \eqref{eq:ms_val}) to potential (or \textit{Oracle}) performances $\mathcal{L}^{**}$ obtained with ideal metrics $\mathcal{L}_{pot}$ (Equation \eqref{eq:oracle_str}) at the model selection task across the search space ($\mathcal{C,B,H}$). Selection is always done on validation data and averaged across $10$ CV folds, whereas the final causal estimation performances are always reported on the test set, \rev{which is repeated in} $10$ train-select-test cycles \rev{per dataset}. Observable metrics $\mathcal{L}_{obs}$ use observational data $\mathcal{D}_{obs}$; Oracle selection and performances use $\mathcal{D}_{pot}$ that includes all potential outcomes.

The setup involves 4 datasets and 4 potential metrics (Section \ref{sec:data}), 5 observable metrics (Section \ref{sec:val_met}) and multiple CATE estimator combinations (7 CATE estimators, 9 base learners; Section \ref{sec:est}) coupled with various hyperparameter sets. Observable metrics are also evaluated using different quality measures (Section \ref{sec:val_mem}). To provide a quantitative measure of \textit{impact} (goal a)) and \textit{effectiveness} (goal b)), all aforementioned performances ($\mathcal{L}_{pot}, \mathcal{L}_{pot}^*, \mathcal{L}^{**}$) were used to calculate probabilities of reaching SotA performance levels (Section~\ref{sec:prob_sota}).

\subsection{Data}\label{sec:data}
We incorporate four benchmark datasets commonly used in the treatment effect estimation literature, which were designed to assess performance of CATE estimators. These datasets are: IHDP, Jobs, Twins and News. Depending on the characteristics and available information as part of the data, different performance metrics $\mathcal{L}_{pot}$ are used. With Jobs, $\mathcal{L}_{pot} \in \{ \epsilon_{ATT}, \mathcal{R}_{pol} \}$ metrics are used; otherwise $\mathcal{L}_{pot} \in \{ \epsilon_{ATE}, PEHE \}$. Each dataset consists of a number of slightly different variations of data, often referred to as \textit{realisations} or simply \textit{iterations}. Crucially, the data iterations of each dataset represent random samples from the true population used to generate the datasets. The average performance over these iterations thus represents a Monte Carlo estimate of the performance was the method applied directly to the population. This is why we take Equation \eqref{eq:oracle_str} to have a `strong Oracle' interpretation. The number of available data iterations vary across datasets ($10$-$1,000$). We stick to $10$ across all four, mostly due to computational reasons, as we find $10$ iterations are already strongly indicative of the overall performance of CATE estimators.

\subsection{Observable Evaluation Metrics}\label{sec:val_met}
This study incorporates multiple observable validation metrics $\mathcal{L}_{obs}$ that can be used for model selection purposes.

\subsubsection{MSE}
The usual MSE that measures aggregate squared error on factual outcomes. Lower is better.
\begin{equation}
    \muRisk = \frac{1}{n}\sum_{i=1}^n(y_i - \hat{\mu}_t(x_i))^2
\end{equation}

\subsubsection{Plugin Validation}
It involves fitting a CATE model on the validation set, and then treating the estimator as a source of (pseudo) ground truth a candidate model trained on the training data is evaluated against. In order to avoid making predictions on the same dataset that was used for training, the usual approach is to perform \textit{cross-fitting}. The idea is similar to \textit{cross-validation}, where data is split into K folds of the same size. The training phase is exactly the same as in CV, but instead of performing evaluation on the validation folds, we make predictions on them. Through this process we obtain $\tau_{plug}$ estimator which can be used as an alternative source of ground truth when calculating various CATE metrics. The usual approach is to use the formula for $PEHE$ but use $\tau_{plug}$ instead of the actual ground truth for individual effects. This gives us the \textit{Plugin PEHE} formula as:
\begin{equation}
    \tauRisk_{plug}^{PEHE} =\sqrt{\frac{1}{n}\sum_{i=1}^{n}( \hat{\tau}(X_i) - \tau_{plug}(X_i))^2}
\end{equation}

Since $\tau_{plug}$ can be treated as simply the source of true CATEs, a \textit{Plugin ATE} formula get also be obtained:
\begin{equation}
    \tauRisk_{plug}^{ATE} =\left \lvert \frac{1}{n} \sum_{i=1}^{n} \hat{\tau}(X_i) - \frac{1}{n} \sum_{i=1}^{n} \tau_{plug}(X_i) \right \rvert
\end{equation}

Both of the above are separate model selection strategies as they provide different feedback signals. The following estimators and base learners are used to obtain $\tau_{plug}$. CATE estimators: S-Learner and T-Learner. Base learners: Decision Tree, Boosted Trees (LightGBM), and Kernel Ridge. Basic model selection is also performed first for base learners. The same hyperparameter search space is used as for the proper CATE estimators. $R^2$ metric is used to select the best combination via 5-fold CV stratified on treatment \textit{T} \citep{kohaviStudyCrossvalidationBootstrap1995}. Once selection is done, a proper learning is performed to obtain $\tau_{plug}$.

\subsubsection{Matching}
As any CATE estimator can be plugged in to $\tauRisk_{plug}$ as $\tau_{plug}$, we also experiment with \textit{matching}. For clarity, we denote it as $\tauRisk_{match}$ to distinguish it from regular plugins.

We obtain $\tau_{plug}$ through \textit{k-Nearest Neighbours} (kNN) algorithm. Data points of each treatment group are stored by a separate kNN instance, and then matching pairs are found using opposite kNN instances. For example, to find counterfactuals for all controls, control units are passed to the kNN instance that stored treated units, resulting in treated units that match the best the control ones. The same process is then repeated for the other treatment group. Overall result is predicted (matched) counterfactuals, which combined with factuals can be used to obtained CATEs, further used as pseudo ground truth.

We use Euclidean distance in the kNN to find matching pairs. Along with the default single-neighbour matching ($k=1$), we also experiment with $k=3$ and $k=5$. In those cases, the resulting counterfactual prediction is an average of k nearest matches inversely weighted by their distance to the query point (closer points have greater influence). Similarly to $\tauRisk_{plug}$, we also investigate the effectiveness of both $PEHE$ and $ATE$ variants.

\subsubsection{R-Score}
We start by defining two nuisance functions $m(X)~=~\mathbb{E}\left [ Y \mid X \right ]$ and $e(X)~=~\mathbb{E}\left [ T \mid X \right ]$, estimates of which are plugged into the \textit{R-Loss} formula $\mathcal{L}(\tau) = \frac{1}{n}\sum_{i=1}^{n}( [Y_i - \hat{m}(X_i)] - [T_i - \hat{e}(X_i)]\tau(X_i) )^2$ as per \citep{nieQuasioracleEstimationHeterogeneous2021}. By inserting any candidate CATE estimator $\hat{\tau}$, the \textit{R-Score}, formulated by \citet{econml}, is then of the following form:

\begin{equation}
    \tauRisk_R = 1 - \frac{\mathcal{L}(\hat{\tau})}{\min_{\tau} \mathcal{L}(\tau)}
\end{equation}

With the denominator being a simple linear CATE model that achieves the lowest loss. Similarly to $R^2$ that penalises constant-valued predictions, $\tauRisk_R$ treats linear CATE model as its baseline and hence penalises predicting constant effects. Higher is better, with a value of $1.0$ being a perfect score. Negative score suggests worse performance than predicting constant effects.

Estimates of the nuisance functions $m(X)$ and $e(X)$ are obtained through \textit{cross-fitting}, same as with $\tauRisk_{plug}$. Consequently, the same base learners are used to obtain $\hat{m}(X)$ and $\hat{e}(X)$, that is, Decision Trees, LightGBM and Kernel Ridge. The only difference being the need to use regressors and classifiers to get $\hat{m}(X)$ and $\hat{e}(X)$ respectively. In addition, basic model selection is performed for both nuisance functions via stratified 5-fold CV \citep{kohaviStudyCrossvalidationBootstrap1995}.

The initial idea of the \textit{R-Loss} comes from \citep{nieQuasioracleEstimationHeterogeneous2021}. They recommend using \textit{Gradient Boosted Trees} (XGBoost), which we do incorporate but in the form of LightGBM. Further extension to a more robust \textit{R-Score} (here $\tauRisk_R$) is credited to \citep{econml}.

\subsection{Estimators}\label{sec:est}
We use the the following \textbf{CATE estimators}: S-Learner (SL), T-Learner (TL), X-Learner (XL) \citep{kunzelMetalearnersEstimatingHeterogeneous2019}, Doubly Robust (DR) \citep{robinsEstimationRegressionCoefficients1994}, Double Machine Learning (DML) \citep{chernozhukovDoubleDebiasedMachine2018}, Inverse Propensity Score Weighting (IPSW) \citep{hiranoEfficientEstimationAverage2003}, Causal Forest (CF) \citep{atheyGeneralizedRandomForests2019a}. Many provided by EconML \citep{econml}. We then combine with the following \textbf{Base learners}: Lasso Lars (L1), Ridge Regression (L2), Decision Trees (DT), Random Forest (RF), Extremely Randomised Trees (ET) and Kernel Ridge (KR) accessed via scikit-learn \citep{scikit-learn}. Gradient Boosting Trees as LightGBM (LGBM) \citep{keLightGBMHighlyEfficient2017} and CatBoost (CB) \citep{prokhorenkovaCatBoostUnbiasedBoosting2018}. Feedforward Neural Networks (NN) realised via TensorFlow \citep{tensorflow2015-whitepaper}.

Each combination of CATE estimators and base learners is treated as a separate model, such as \textit{SL-L1} or \textit{DR-RF}. We combine all learners to obtain the experimental setup. The only exceptions are NNs, combined only with SL and TL for computational reasons, and CF (standalone estimator, no choice of base learners). All CATE estimators, base learners and hyperparameters refer to search spaces $\mathcal{C}$, $\mathcal{B}$ and $\mathcal{H}$ respectively in Equation \eqref{eq:search}. Following standard industry practice, we perform 10-fold cross-validation~\citep{kohaviStudyCrossvalidationBootstrap1995}.

\subsection{Validation of Observable Evaluation Metrics}\label{sec:val_mem}
In order to investigate the effectiveness of observable evaluation metrics, we incorporate the following quality measures. First, we follow Equation \eqref{eq:ms_val}, which measures the performance of $\mathcal{L}_{obs}$ as the performance of the model $\mathcal{P}^*_{obs}$ but evaluated using $\mathcal{L}_{pot}$. All observable metrics are evaluated against all potential metrics available for the dataset. \textit{Oracle} performances $\mathcal{L}^{**}$ (Equation \eqref{eq:oracle_str}) are also collected for all potential metrics and compared to those model performances achieved via observable metrics. While \textit{Oracle} performances are not accessible when dealing with real datasets (lack of ground truth), it is certainly a useful tool for this study to measure how much biased different observable metrics can be. Moreover, using this method can shed some more light on real capabilities of various CATE estimators, showing what they can potentially achieve if model selection choices are optimised for the ideal target $\mathcal{L}_{pot}$.

\rev{
\subsection{Probability of State-of-the-Art Performance}\label{sec:prob_sota}
Because the main interest of this study is to assess the impact of different hyperparameter values and metrics on estimation performance, we introduce a measure of success that objectively captures this quality: the probability of a particular hyperparameter choice achieving SotA performance as measured by a given metric. Calculating this probability involves randomly generating repeated iterations of the dataset from the true population model. Such iterations are available to us in IHDP, Jobs, Twins and News datasets. The average of these iterations is an estimate of the true probability for a given estimator-learner-hyperparameter combination. Furthermore, the variation in the performance level across iterations allows us to quantify the Monte Carlo error on this estimate as follows.

First, defining $I(E)=1$ if event $E$ is true and $I(E)=0$ if it is false, we calculate $I(\mathcal{L}^*_i \leq \mathcal{L}_{SotA})$ for each iteration $i$, that is, whether performance $\mathcal{L}^*_i$ reaches or exceeds SotA level $\mathcal{L}_{SotA}$.  We then assess overall performance using ${\hat p}$, the proportion of times the combination exceeds SotA, where
\begin{equation}\label{eq:p_sota}
{\hat p} = \frac{1}{N} \sum_{i=1}^N I(\mathcal{L}^*_i \leq \mathcal{L}_{SotA}) \sim \mathcal{N}(p, (1-p)p/N).
\end{equation}
In other words, ${\hat p}$ is normally distributed (across hypothetical replications of our datasets each based on $N$ different iterations from the same population model) with mean $p$ - the true probability of achieving SotA performance - and variance $(1-p)p/N$. The Monte Carlo error of ${\hat p}$ can then be quantified using the 95\% confidence interval $\left({\hat p}\pm 1.96\sqrt{(1-{\hat p}){\hat p}/N}\right)$.

Note that each dataset consists of $N{=}10$ iterations and so the distributional result in Equation~\eqref{eq:p_sota} is an approximation and, typically, one only taken to be reliable if true $Np>5$. It is possible that $Np<5$ if $N{=}10$, in which case we take the interval above to be a useful first-order approximation of the true confidence interval and, as such, an indicative measure of Monte Carlo error. However, because $N{=}40$ when we merge probabilities across all four datasets, $Np>5$ is more plausible and so the 95\% coverage level of the confidence intervals is reasonably taken to be accurate.
}

\section{Results and Discussion}\label{sec:results}

\rev{
To obtain SotA probabilities as described in Section~\ref{sec:prob_sota}, we gathered results from the recent literature on causal effect estimation. These include TARNet and CFR-WASS \citep{shalitEstimatingIndividualTreatment2017}, SITE \citep{yaoRepresentationLearningTreatment2018}, GANITE \citep{yoonGANITEEstimationIndividualized2018}, CEVAE \citep{louizosCausalEffectInference2017a}, as well as BLR, BNN-4-0 and BNN-2-2 \citep{johanssonLearningRepresentationsCounterfactual2016}. Together, these form a range of values that here we consider as SotA performance levels, which we used to obtain probabilities for all considered CATE estimators and base learners across all four datasets. Note that $\mathcal{L}_{SotA}$ in Equation~\eqref{eq:p_sota} changes depending on the dataset and the potential metric used for evaluation on the test set. Furthermore, working with probabilities (Equation~\eqref{eq:p_sota}) notably allows us to aggregate the results across all datasets, resulting in a more unified view. The only exception is Twins dataset for which no results with respect to $\epsilon_{ATE}$ metric are available, hence no $\epsilon_{ATE}$ SotA for Twins. However, we do categorise obtained probabilities into those achieved on average and individualised treatment effect estimation tasks, for which we use $\{\epsilon_{ATE}, \epsilon_{ATT} \}$ and $\{PEHE, \mathcal{R}_{pol} \}$ potential metrics respectively.

The main results presented in this section are supplemented with additional material available in Appendix~\ref{app:results}. All experiments can be replicated using our code that is available online.\footnote{\url{https://github.com/misoc-mml/hyperparam-sensitivity}}
}

\begin{figure}[tb]
    \centering
    \includegraphics[width=0.90\textwidth]{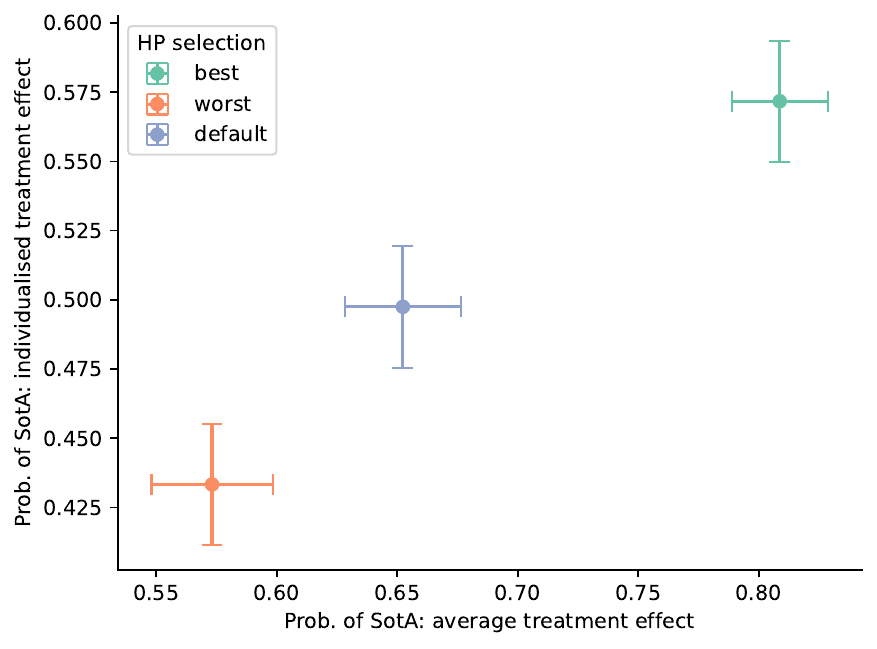}
    \caption{\rev{Probability of reaching SotA performance levels (higher is better) depending on the quality of hyperparameters (colours). The results are aggregated across all CATE estimators and datasets. Error bars: 95\% confidence intervals. X-axis: $\epsilon_{ATE}$ and $\epsilon_{ATT}$. Y-axis: $PEHE$ and $\mathcal{R}_{pol}$. Interpretation: default hyperparameters are not optimal; the Oracle performances, achieved via the best hyperparameter values and selected with potential metrics, have significantly higher probability of reaching SotA.}}
    \label{fig:worst_def_oracle}
\end{figure}

Our first goal is to analyse how the quality of hyperparameters impacts effect estimation performance across different causal estimators. For each CATE estimator-base learner combination ($\mathcal{C}_c, \mathcal{B}_b$), and across all hyperparameters $\mathcal{H}$, we select three types of hyperparameters: \textit{default} as defined by code packages, those that achieve the \textit{worst} potential metrics values (max $\mathcal{L}_{pot}$), and those that attain the \textit{best} potential metrics values (min $\mathcal{L}_{pot}$, effectively $\mathcal{L}^{**}$, or \textit{Oracle}). \rev{We then calculate probabilities of reaching SotA for all estimator-learner combinations and each type of hyperparameters as defined above. The last step involves merging all probabilities across estimators, learners and datasets ($N$ in Equation~\eqref{eq:p_sota} is adjusted accordingly).} Figure~\ref{fig:worst_def_oracle} depicts obtained results. Clearly, default hyperparameters are not optimal (blue), and the best-performing hyperparameters (green), \rev{significantly increase the probability of reaching SotA in both estimation tasks (x-axis: $0.65(0.02) \rightarrow 0.81(0.02)$, y-axis: $0.50(0.02) \rightarrow 0.57(0.02)$).}

\begin{figure}[tb]
    \centering
    \includegraphics[width=0.95\textwidth]{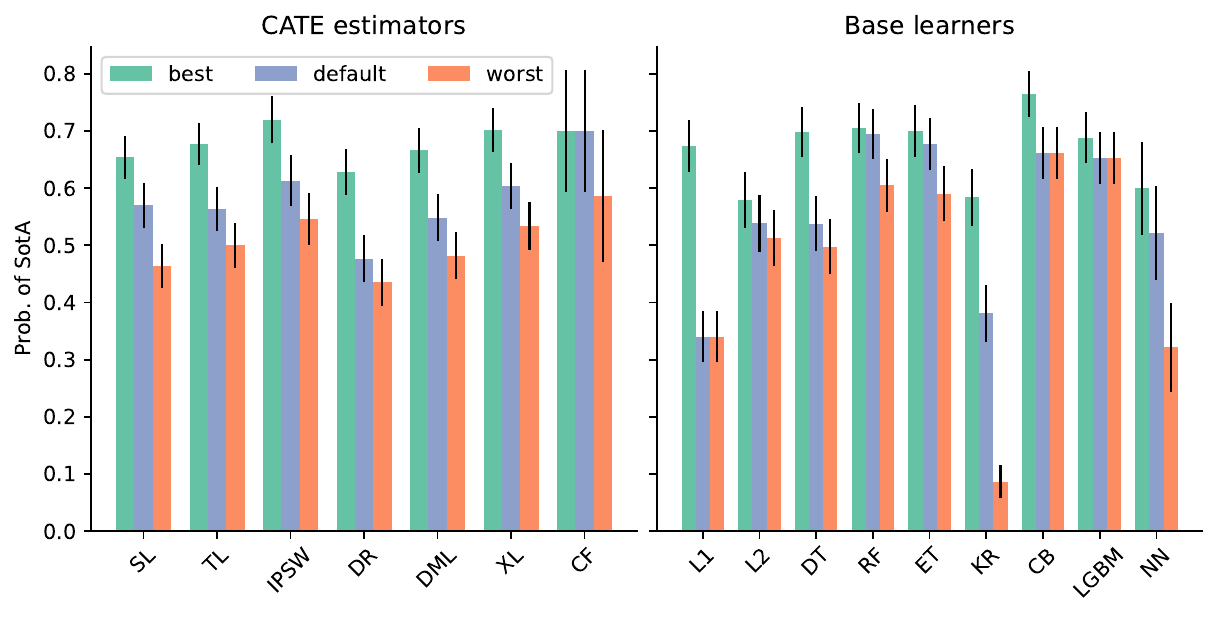}
    \caption{\rev{Probability of reaching SotA performance levels (higher is better) by individual CATE estimators (left) and base learners (right) depending on the quality of hyperparameters (colours). The results are aggregated across all datasets and potential metrics. Error bars: 95\% confidence intervals. Interpretation: regardless of estimators and learners, the best hyperparameters provide mild or significant improvements in probability of reaching SotA as compared to default hyperparameters (with some exceptions).}}
    \label{fig:est_bl}
\end{figure}

In our further analysis, we take a closer look at how estimation performance varies across different causal estimators \rev{and base learners} when picking \rev{different types of} hyperparameters. This is to investigate \rev{which of the two types of choices are more important to achieve strong performance: selection of estimators and learners, or selection of hyperparameters. To this end, we perform an analysis similar to that in Figure~\ref{fig:worst_def_oracle} but make two changes: a) we group probabilities by CATE estimators and base learners, and b) we merge probabilities across the average and individual estimation tasks.} The results are presented in Figure \ref{fig:est_bl}. \rev{In terms of CATE estimators (left subfigure), the best hyperparameters significantly increase the probability of SotA over default values (typical gains within $8{-}12\%$) across almost all estimators, with CF being the only exception. As for base learners (right subfigure), all show at least mild gains with the best hyperparameters over the defaults, with L1, DT, KR and CB showing particularly significant improvements (gains between $10{-}33\%$ in those four cases). Three of the ensemble methods (RF, ET, LGBM) exhibit particularly robust default hyperparameters, which likely explains similarly strong default performance of the CF CATE estimator due to its ensembling nature. Overall, the best hyperparameters provide at least mild or significant improvements in SotA probability across most estimators and learners, making hyperparameter selection a possibly more important task than selecting estimators and learners (i.e. tuning a single model is likely to yield better results than selecting from multiple models without tuning).}

\begin{figure}[tb]
    \centering
    \includegraphics[width=0.90\textwidth]{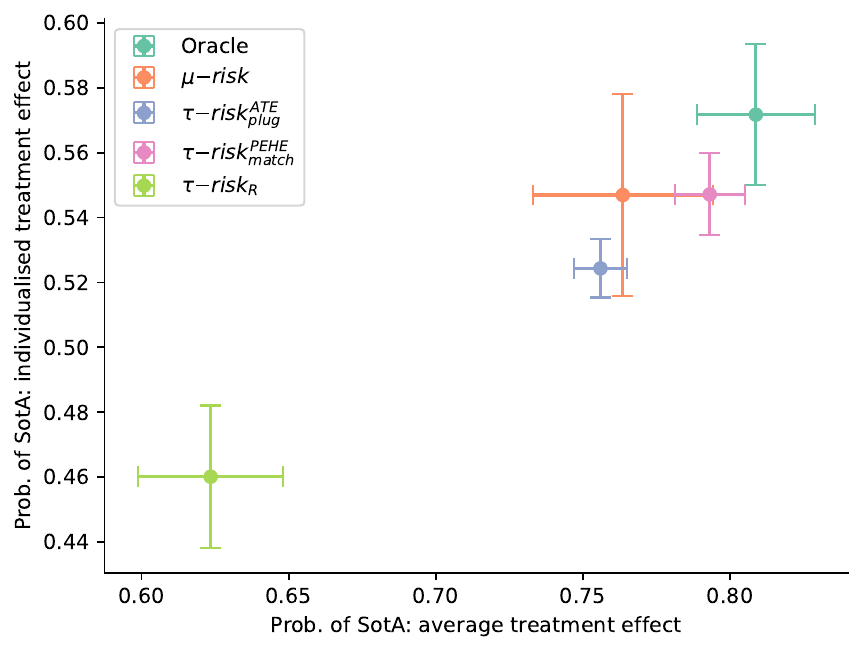}
    \caption{\rev{Probability of reaching SotA performance levels (higher is better) depending on used model selection metrics (colours). The results are aggregated across all CATE estimators and datasets. Error bars: 95\% confidence intervals. X-axis: $\epsilon_{ATE}$ and $\epsilon_{ATT}$. Y-axis: $PEHE$ and $\mathcal{R}_{pol}$. Only the best performing observable metrics are included for readability. The metrics presented with different colours were used to select candidates on the validation data, and then evaluated on the test data using potential metrics (axes x and y). Interpretation: the choice of a metric can seriously impact SotA probability and by extension the final estimation performance.}}
    \label{fig:metrics}
\end{figure}

Our results presented so far show that estimators equipped with the right hyperparameters can \rev{reach SotA performance levels with high probability}. However, finding those well performing hyperparameters in a data-driven way requires an observable metric that can be used in the process of hyperparameter tuning. In Figures \ref{fig:worst_def_oracle} and \ref{fig:est_bl}, we used potential metrics to select the best hyperparameters to show true capabilities of common causal estimators, but those metrics cannot be used \rev{for} tuning in practice as they require access to true CATEs (inaccessible in real life scenarios). For this reason, the next step is to investigate how \rev{various observable metrics influence the probability of reaching SotA. The analysis is again similar to that in Figure~\ref{fig:worst_def_oracle}, where SotA probabilities are merged across all estimators, learners and datasets, but here we test multiple observable metrics to select the best hyperparameters on the validation data. We further compare the SotA probabilities obtained with observable metrics to those achieved with the ideal potential ones (i.e. Oracle). The results presented in Figure~\ref{fig:metrics} include only the best performing observable metrics in their respective categories for increased readability. The first, and by far the most critical for practice observation we make, is that SotA probability can vary substantially among metrics. This shows the choice of a selection metric can seriously impact the final estimation performance. The degree of variability is also considerable with some metrics, such as $\muRisk$ and $\tauRisk_R$, making them potentially risky choices in practical applications. $\tauRisk_R$ particularly seems to be performing very differently as compared to the rest of the metrics and other recent studies~\citep{doutreligne2025select}. A possible explanation is that certain metrics may be more compatible as measures when combined with specific estimators/learners~\citep{curth2023search}, which is not necessarily reflective of the metric's true capabilities but rather our choice of estimators. Furthermore, MSE ($\muRisk$), a go-to tool in ML, while being in close proximity to the Oracle, may produce inconsistent results. The second main observation is that some observable metrics are within close proximity to the Oracle, which specifically applies to $\muRisk$ and $\tauRisk^{PEHE}_{match}$. The latter choice looks particularly promising due to its low variability (95\% CI $\approx 0.01$). This shows a hopeful picture that the very best of hyperparameters are perhaps within reach when using the right tools. Though the question may still remain as to which metric is the best for a particular task at hand~\citep{curth2023search}. }

\subsection{Limitations}\label{sec:limits}
This study is limited in a few ways. First, we make the usual set of assumptions about the data, that is, \textit{SUTVA} and \textit{strong ignorability}, which may or may not hold in practice. Secondly, the experimental setup has some shortcomings, mostly due to computational reasons. While an effort was put to explore relatively wide search spaces to make our findings as general as possible, a practical balance had to be found to ensure computational feasibility. This applies to explored hyperparameters of base learners as well as combinations of base learners in CATE estimators. More concretely, if a CATE estimator uses multiple base learners, only one type of learners is explored at a time, never mixing multiple types of learners simultaneously. For instance, in the case of a \textit{T-Learner} with two internal base learners per treatment arm, if the goal is to combine it with Decision Trees then both internal regressors will be Decision Trees. If we explore it with Random Forests, both regressors are then Random Forests. And so on. As some CATE estimators incorporate as many as 5 base learners (e.g. \textit{X-Learner}), we further cut down the search space by assigning the same hyperparameter values for all base learners while exploring different variations of hyperparameters. This is possible because all base learners are always of the same type within a CATE estimator. The setup is also limited by our choice of CATE estimators, base learners, model evaluation metrics and datasets. However, regardless of those shortcomings, we believe the proposed setup is general enough to support our findings, and extending it would unlikely alter the main observations in a fundamental way while unnecessarily increasing computational burden significantly.

\rev{While we mostly focus throughout on population-level performances of the estimator-hyperparameter combinations, we also include between-iteration variability to compare stability of estimators and metrics. Note, however, that in our analysis we do not distinguish between iterations of different datasets (i.e. all iterations across all four datasets are merged together) as we do not seek to study between-dataset variability specifically. Moreover,} this approach, which leans heavily on the availability of iterations, is not a luxury that analysts have in practice, but an artefact of our focus on partially synthetic benchmark datasets for causal methods (e.g. \citet{hillBayesianNonparametricModeling2011}). In practice, the analyst would have the equivalent of a single iteration where population-level quantities are not accessible. In such cases, the analyst could use bootstrapping to make inferences about the true population value, but we do not investigate the performance of the bootstrap in this paper (this is not the focus of our study and would increase the already substantial computational burden of our setup).

Our \rev{strong} interest in general trends is also reflected in not \rev{specifically} analysing how robust are individual estimators \rev{and base learners} to hyperparameter choices\rev{, despite showing their between-iteration variability.} We instead focus on the general observation that hyperparameters selected with Oracles \rev{significantly increase} SotA \rev{probability} \textbf{across \rev{almost} all estimators}. This general viewpoint continues when we conclude that in the causal inference pipeline, hyperparameters and metrics are the main drivers of strong predictive performances, leaving the choice of estimators as not so influential and hence less important. Thus, as our results point to estimator-agnostic conclusions, we deem any further analysis into differences among estimators as unnecessary and out of scope of this work.

\section{Conclusion}\label{sec:conclusion}
This study provided an empirical evidence that default hyperparameters are not optimal; the best hyperparameters, selected with ideal metrics, \rev{can significantly increase the probability of reaching SotA performance levels} in causal effect estimation. In most cases, this holds true across different types of causal estimators and ML base learners, encouraging to prioritise the selection of good hyperparameters over the choice of estimators and learners. Another implication is that by only changing hyperparameters, an arbitrary performance can be achieved (excellent or poor). This places a big question mark on published benchmarks that test various estimators. If hyperparameters account for such a big part of the estimation performance, should claimed performances be credited to the method itself or well-identified hyperparameters? The fact that studies often use different metrics to optimise hyperparameters only adds complexity to this issue. Furthermore, we also show the \rev{SotA probability achieved with different metrics} varies significantly. As a consequence, the choice of model evaluation metric highly impacts resulting estimation performance. In addition, we show that \rev{some} observable metrics (feasible with observational data) \rev{can achieve the same level of SotA performance} as those \rev{achieved using} ideal (but \rev{generally} inaccessible) metrics. \rev{This brings hope in using the metrics in applied settings, though choosing the right metric for the task at hand remains an open challenge~\citep{curth2023search}.} All these issues encourage to consider the development of more robust model selection methods.

Through substantial empirical evidence, we make a strong case that more focus on metrics is needed for causal learning problems. In this, we connect with the theoretical work on the validation of causal models by \citet{alaaValidatingCausalInference2019}.\footnote{Not included in this work due to unavailable public implementation of the method, and us focusing on metrics widely available to practitioners.} In their framework, oMSE is a special case of plug-in loss in which the counterfactuals are `synthesised' through elements of the candidate causal model (their equation (6)), and pMSE a special case of expected loss under the true causal model used to generate the observed data (their equation (5)). They propose a targeting procedure based on theoretically derived influence functions to update the plug-in estimate to lie closer to the solution which would have been obtained had pMSE been available (see their Theorem 1). The approach implemented in their paper (based on a first-order Von Mises expansion as set out in their Theorem 2) will work well in terms of choosing the best model if the plug-in estimate lies inside a tight neighbourhood of the true value, but our results indicate that this cannot be guaranteed to hold. We argue strongly, therefore, that more attention to overcoming this problem is needed from theorists and methodologists (e.g. implementing Alaa and Schaar’s influence function approach for higher-order, and thus more accurate, Von Mises expansions) because this is of vital importance for practice.

An overall takeaway is that, contrary to common belief, popular causal estimators can provide excellent estimation performance given the right hyperparameters, but \rev{hyperparameter selection is challenging. The use of MSE as a metric, a popular choice in ML, is rather discouraged due to its high variability that will lead to inconsistent results. Some metrics (matching-based naive plug-in, R-Loss) did not perform as expected~\citep{doutreligne2025select}, possibly highlighting the sensitivity of evaluation measures to the choice of estimators and learners~\citep{curth2023search}. Thus, until more robust observable metrics are developed, our practical recommendation is to leverage} highly interpretable models, such as linear regression or decision trees. \rev{These} provide intuitive hyperparameters that can be manually tuned to the problem at hand by domain experts through them ingraining their prior knowledge into the model via hyperparameters.

In terms of future work, given excellent performances achieved here with already established estimators, we argue new causal estimators are not needed for meaningful progress in causal effect estimation. Rather a more thorough understanding of causal model selection and hyperparameter tuning is required, as once recognised in ML \citep{bergstraAlgorithmsHyperParameterOptimization2011}, that will eventually unlock those great performances we showed are possible but \rev{with consistency and} on real-life observational datasets. Recent studies indeed push the frontier in this area (e.g. \citet{alaaValidatingCausalInference2019,saitoCounterfactualCrossValidationStable2020a,nieQuasioracleEstimationHeterogeneous2021}), but many of the proposed model selection methods involve another layer of learning, on top of causal estimation, which is again subject to hyperparameter tuning, making them very unstable in practice. Perhaps there are alternative ways to be explored of improving causal hyperparameter tuning without relying on learning methods. Moreover, as the number and complexity of causal model selection approaches increases, new tools and frameworks might be needed to facilitate the use of the latest selection methods and help practitioners in the complex task of model evaluation in the causal setting \citep{shimoniEvaluationToolkitGuide2019}. Finally, due to the importance of hyperparameters in the estimation performance and inconsistent practices in published benchmarks, some form of standardisation in model tuning across the field of causal inference might be necessary to overcome this issue in the future. One possibility is to develop a public benchmarking platform that would test and tune all types of causal estimators equally thoroughly, following ideas from the causal discovery community \citep{rios2021benchpress}.

\subsubsection*{Acknowledgments}
We would like to thank Ehud Karavani for invaluable feedback and suggestions. All three authors (DM, SS and PC) were supported by the ESRC Research Centre on Micro-Social Change (MiSoC) - ES/S012486/1. This research was supported in part through computational resources provided by the Business and Local Government Data Research Centre BLG DRC (ES/S007156/1) funded by the Economic and Social Research Council (ESRC). DM was also supported by the Engineering and Physical Sciences Research Council AI Hub for Causality in Healthcare AI with Real Data (CHAI) - EP/Y028856/1.

\bibliography{tmlr}
\bibliographystyle{tmlr}

\newpage
\appendix

\section{Datasets Details}\label{app:datasets}

\subsection{Metrics}\label{app:metrics}
Error on ATE is the absolute difference between true and predicted ATEs.
\begin{equation}\label{eq:e_ate}
    \epsilon_{ATE}=\left \lvert \frac{1}{n} \sum_{i=1}^{n} \hat{\tau}(x_i) - \frac{1}{n} \sum_{i=1}^{n} \tau(x_i) \right \rvert
\end{equation}

Precision in Estimation of Heterogeneous Effect (PEHE) is the root mean squared error between true and predicted CATEs.
\begin{equation}\label{eq:pehe}
    PEHE=\sqrt{\frac{1}{n}\sum_{i=1}^{n}( \hat{\tau}(x_i) - \tau(x_i))^2}
\end{equation}

Given a set of treated subjects $T$ that are part of sample $E$ coming from an experimental study, and a set of control group $C$, the true Average Treatment effect on the Treated (ATT) and its error are defined as:
\begin{equation}
    ATT = \frac{1}{\lvert T \rvert}\sum_{i \in T} \mathcal{Y}^{(i)} - \frac{1}{\lvert C \cap E \rvert}\sum_{i \in C \cap E} \mathcal{Y}^{(i)}
\end{equation}
\begin{equation}\label{eq:e_att}
    \epsilon_{ATT} = \left \lvert ATT - \frac{1}{\lvert T \rvert} \sum_{i \in T} \hat{\tau}(x_i) \right \rvert
\end{equation}

Policy risk can be defined as:
\begin{multline}
    \mathcal{R}_{pol} = 1 - (\mathbb{E}\left [ \mathcal{Y}_1 
\mid \pi(x)=1 \right ] \mathcal{P}(\pi(x)=1)  \\
    + \mathbb{E}\left [ \mathcal{Y}_0 \mid \pi(x)=0 \right ] \mathcal{P}(\pi(x)=0))
\end{multline}\label{eq:r_pol}
Where $\mathbb{E}[.]$ denotes mathematical expectation and policy $\pi$ becomes $\pi(x)=1$ if $\hat{\tau}(x) > 0$; $\pi(x)=0$ otherwise. Note the $\mathcal{R}_{pol}$ defined in this way is always estimated due to its dependency on estimated causal effects $\hat{\tau}(x)$, regardless whether the predictions are made on the validation or the test set.

\subsection{Descriptions}
\textbf{IHDP} was introduced by \citet{hillBayesianNonparametricModeling2011}, based on Infant Health Development Program clinical trial \citep{brooks-gunnEffectsEarlyIntervention1992}. The goal is to estimate the effect of specialised childcare services on infants' cognitive test scores conducted after some time. All outcomes (including counterfactuals) are simulated using real covariates from the experiment, though only factual outcomes are exposed to CATE estimators to maintain observational data setting. Metrics are $\epsilon_{ATE}$ and $PEHE$. Training/test splits are in 90/10 ratio. There are $25$ background features and $747$ rows of data ($139$ treated and $608$ controls).

\textbf{Jobs}, proposed by \citet{a.smithDoesMatchingOvercome2005}, is a combination of experimental \citep{lalondeEvaluatingEconometricEvaluations1986} and observational data \citep{dehejiaPropensityScoreMatchingMethods2002}. Data variables are basic characteristics of unemployed at the time participants ($17$ features $X$), whether they participated in a job training ($T$), and their employment status later on ($Y$). Metrics are $\epsilon_{ATT}$ and $\mathcal{R}_{pol}$. Training/test splits follow 80/20 ratio. There are $3,212$ units ($297$ treated and $2,915$ controls).

\textbf{Twins} was compiled from twin births in the US between 1989-1991 \citep{almondCostsLowBirth2005}. We incorporate the version pre-processed by \citet{louizosCausalEffectInference2017a}, which consists of only same-sex units with weight below $2kg$. The task is to estimate the effect of higher weight (heavier twin) on mortality in a span of the first year of life. In addition, one of the twins is always hidden to simulate observational setting. Note the usual metric of choice in the literature for this dataset is $AUC$ (area under the ROC curve), which is suitable only for CATE estimators providing outcome predictions (both $\hat{y}_0$ and $\hat{y}_1$). As our experimental setup includes many CATE estimators that do not provide predicted outcomes, we switch back to the usual $\epsilon_{ATE}$ and $PEHE$ metrics. Training/test splits are 80/20. $194$ input features and $11,984$ units (equal split between treatment arms).

\textbf{News}, introduced by \citet{johanssonLearningRepresentationsCounterfactual2016}, is a database of news articles stored as word counts according to predefined dictionary. The task is effect estimation of a device type (mobile or desktop) used to read the article on (simulated) user experience. Training/test splits are 90/10. $3,477$ background covariates and $5,000$ rows ($2,289$/$2,711$ treated and controls respectively).

\newpage
\section{Hyperparameters}\label{app:hyper}
The hyperparameters and corresponding ranges of values that were explored as part of using various base learners are listed in Table \ref{tab:hyper}. Moreover, whenever ensemble learners were used, such as RF, ET, CB, LGBM and CF, the number of inner learners (n\_estimators) was set to $1,000$. When it comes to NNs, we used \textit{relu} activations, $0.25$ dropout, $0.01$ L2 regularisation in the output layer, $10,000$ optimisation steps (not epochs) and Adam optimiser \citep{kingmaAdamMethodStochastic2017}. All hyperparameters refer to the search space $\mathcal{H}$ in Equation \eqref{eq:search}.

\begin{table}[!htb]
    \caption{Hyperparameter search spaces defined per base learner. *Default hyperparameter values.}
    \label{tab:hyper}
    \centering
    \begin{tabular}{lrl}
\toprule
   base learner & hyperparameter & values \\
\midrule
    OLS (L1 and L2) & $alpha$ & $\{.001, .01, .1, .5, 1^*, 2, 10, 20\}$ \\
    {} & $max\_iter$ & $\{1000^*, 10000\}$ \\
    \midrule
    DT, RF, ET and CF & $max\_depth$ & $\{ 2, 3, 4, 5, 6, 7, 8, 9, 10, 15, 20^* \}$ \\
    {} & $min\_samples\_leaf$ & $\{ .01, .02, .03, .04, .05, 1^*, 2, 3, 4, 5, 6, 7, 8, 9 \}$ \\
    \midrule
    KR & $alpha$ & $\{ .001, .01, .1, 1^* \}$ \\
    {} & $gamma$ & $\{ .01, .1, 1^*, 10, 100 \}$ \\
    {} & $kernel$ & $\{ rbf, poly^* \}$ \\
    {} & $degree$ & $\{ 2, 3^*, 4 \}$ \\
    \midrule
    CB & $depth$ & $\{ 5, 6, 7, 8, 9, 10^* \}$ \\
    {} & $l2\_leaf\_reg$ & $\{ 1^*, 3, 10, 100 \}$ \\
    \midrule
    LGBM & $max\_depth$ & $\{ 5, 6, 7, 8, 9, 10^* \}$ \\
    {} & $reg\_lambda$ & $\{ 0, .1^*, 1, 5, 10 \}$ \\
    \midrule
    NN & $hidden\_layers$ & $\{ 1, 2 \}$ \\
    {} & $hidden\_units$ & $\{ 4, 8, 16, 32, 64, 128 \}$ \\
    {} & $learning\_rate$ & $\{ .0001, .001 \}$ \\
    {} & $batch\_size$ & $\{ 32, 64, 128, full \}$ \\
\bottomrule
\end{tabular}
\end{table}

\newpage
\section{Supplemental Results}\label{app:results}

\begin{figure}[htb]
    \centering
    \includegraphics[width=0.9\linewidth]{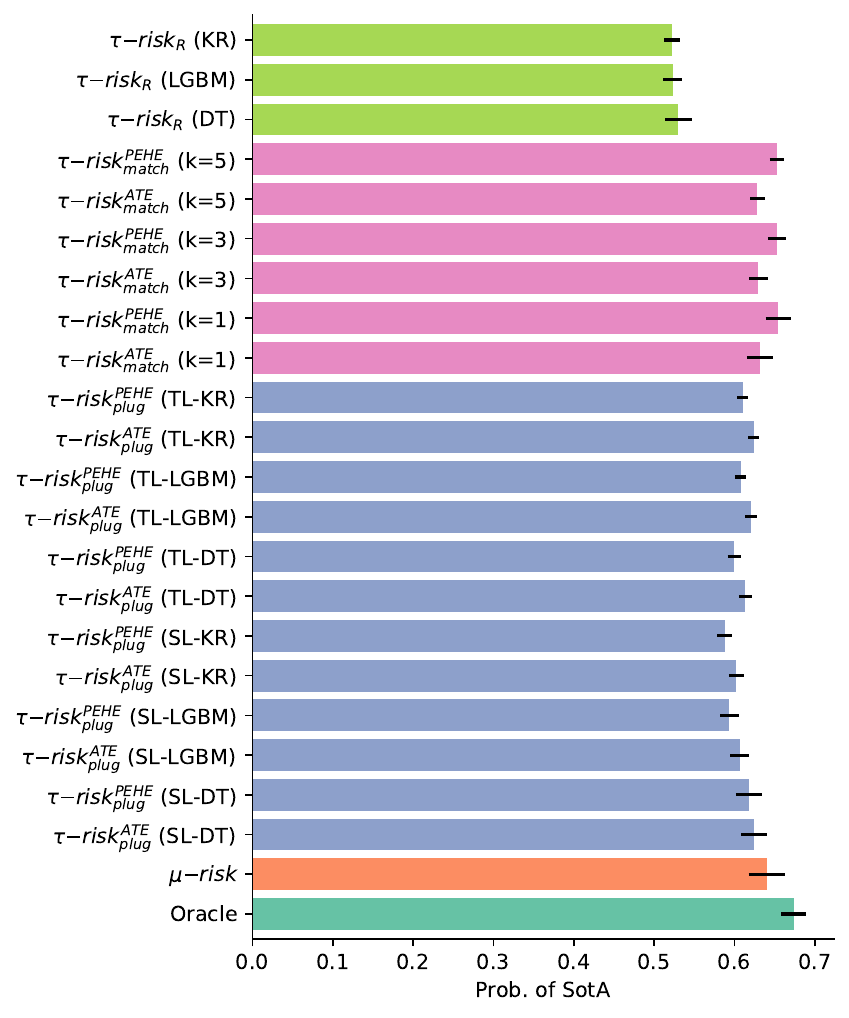}
    \caption{\rev{Probability of reaching SotA performance levels (higher is better) depending on used model selection metrics (y-axis). The results are aggregated across all CATE estimators, datasets and metrics. Error bars: 95\% confidence intervals. The metrics presented along the y-axis were used to select candidates on the validation data, and then evaluated on the test data using potential metrics.}}
    \label{fig:c1}
\end{figure}

\rev{
\subsection{Results for All Metrics}
Figure~\ref{fig:c1} supplements Figure~\ref{fig:metrics} from the main content of the article by showing the probabilities achieved with all metrics and their variants.}

\newpage
\subsection{Alternative Results}\label{app:full_results}
\begin{figure}[htb]
    \centering
    \includegraphics[width=0.90\textwidth]{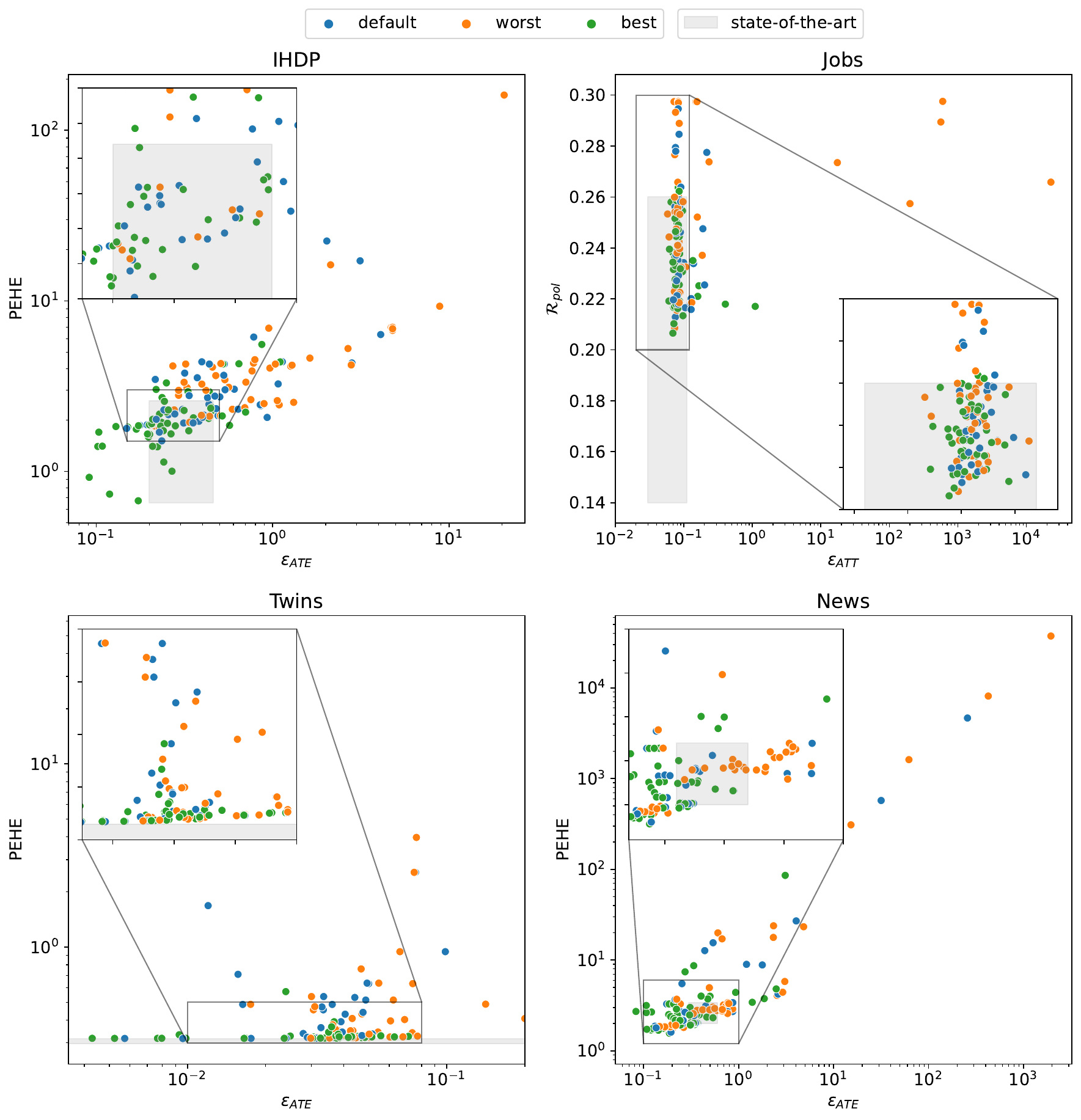}
    \caption{Performance of CATE estimators with three different types of quality of hyperparameters across all four datasets. Each data point represents the mean across dataset iterations. Lower is better applies to all metrics, thus the closer to bottom-left corner, the better. Note the logarithmic scale for all axes to aid visual presentation except policy risk in Jobs. Each of the potential metrics (see axis labels) was used twice here: once to select the best/worst hyperparameter values per estimator on the validation data, and then to evaluate their performance on the test set (shown here).}
    \label{fig:c2}
\end{figure}

\begin{figure}[htb]
    \centering
    \includegraphics[width=0.90\textwidth]{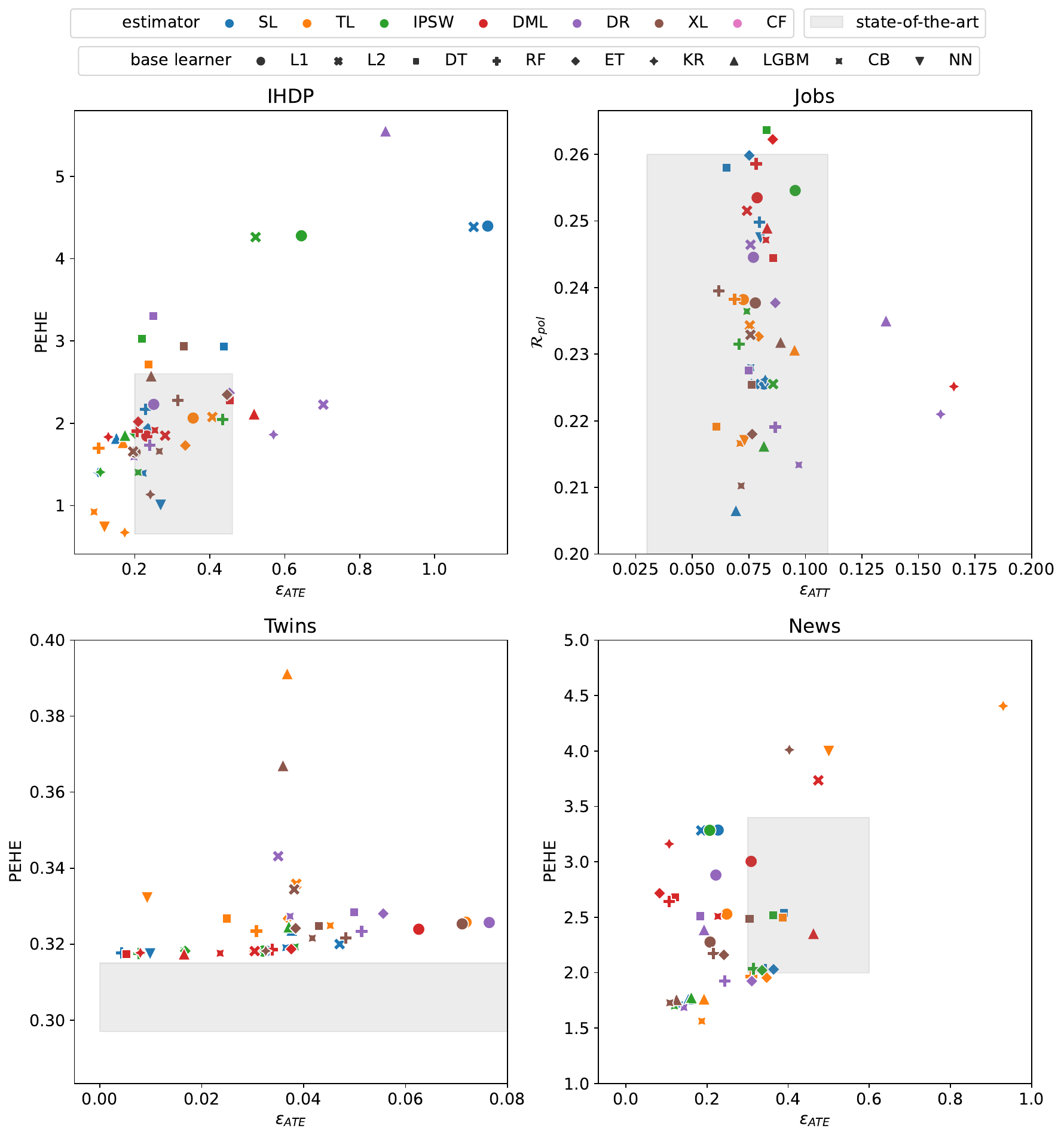}
    \caption{Oracle performances of CATE estimators across all four datasets, grouped by types of causal estimators and base learners. Each data point represents the mean across dataset iterations. Lower is better applies to all metrics, thus the closer to bottom-left corner, the better. Each of the potential metrics (see axis labels) was used twice here: once to select the best hyperparameter values per estimator on the validation data, and then to evaluate their performance on the test set (shown here).}
    \label{fig:c3}
\end{figure}

\begin{figure}[htb]
    \centering
    \includegraphics[width=0.90\textwidth]{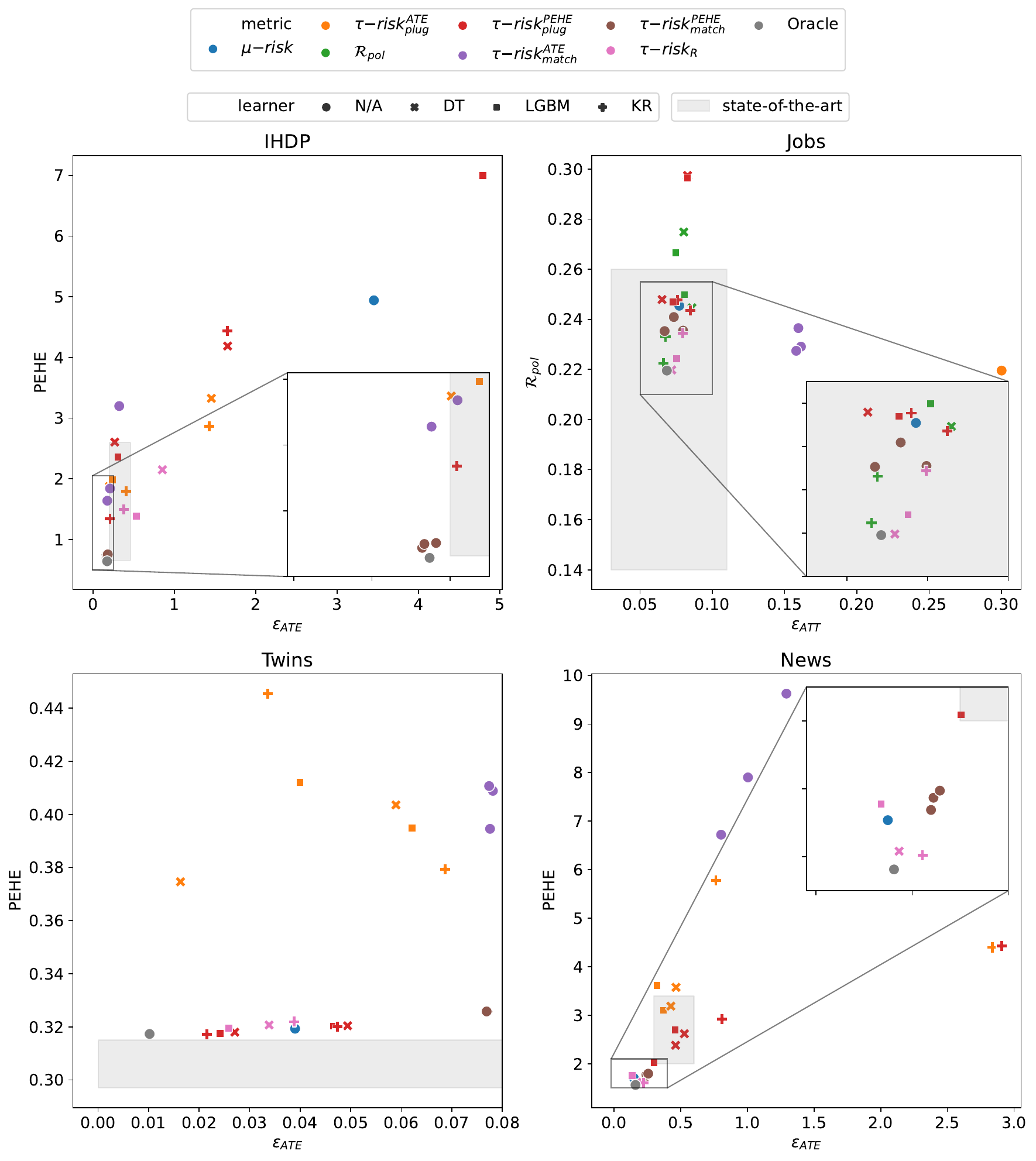}
    \caption{Performance of model selection metrics across all four datasets. Each data point represents the mean across dataset iterations. Lower is better applies to all metrics, thus the closer to bottom-left corner, the better. Each metric (see legend) was used to select the best candidate model on the validation data, including the potential metrics (see axis labels). Then, the potential metrics were used to measure performances of selected candidates on the test set (shown here). The Oracle points represent the use of the potential metrics to select candidates on the validation data, and then evaluated on the test set. When using a potential metric to select on the validation data, the same metric is then used to measure the test performance. Note that some combinations of colours and shapes may appear more than once. This is the case for matching-based ($k\in\{1,3,5\}$) and plugin-based (S- and T-Learning) metrics due to multiple implementation variants explored here.}
    \label{fig:c4}
\end{figure}

\rev{Figures~\ref{fig:c2}-\ref{fig:c4} provide an alternative view on CATE performances and observable metrics as compared to Figures~\ref{fig:worst_def_oracle}-\ref{fig:metrics}. Here, we focus on average CATE estimation performances instead of probabilities and compare them to SotA (see gray areas).}


\clearpage
\subsection{Alternative Oracle}\label{app:test_oracle}

We also explore an alternative version of the Oracle defined in Section \ref{sec:oracle} that performs selection directly on the test set. This is because selecting candidate models on validation data, even with potential metrics, does not guarantee selecting the best possible performances on the test task. As a result, the true estimation capabilities of causal estimators are not demonstrated. We work around this issue by performing Oracle selection directly on the test set. Consequently, the Oracle equations need to be updated to the following form:
\begin{equation}\label{eq:app1}
m^{***} = \argmin_{m=1,...,M}{\mathcal{L}_{pot}(\mathcal{P}_m, \mathcal{D}^{te}_{pot})}
\end{equation}
\begin{equation}\label{eq:app2}
    Oracle_{weak} = \mathcal{L}^{***}_{pot} = \mathcal{L}_{pot}(\mathcal{P}^{***}_{pot}, \mathcal{D}^{te}_{pot})
\end{equation}
\begin{equation}\label{eq:app3}
    Oracle_{strong} = \mathcal{L}^{***}_{pot} = \frac{1}{n} \sum_i^n \mathcal{L}_{pot}(\mathcal{P}^{***}_{pot}(i), \mathcal{D}^{te}_{pot}(i))
\end{equation}

\begin{figure}[htb]
    \centering
    \includegraphics[width=0.90\textwidth]{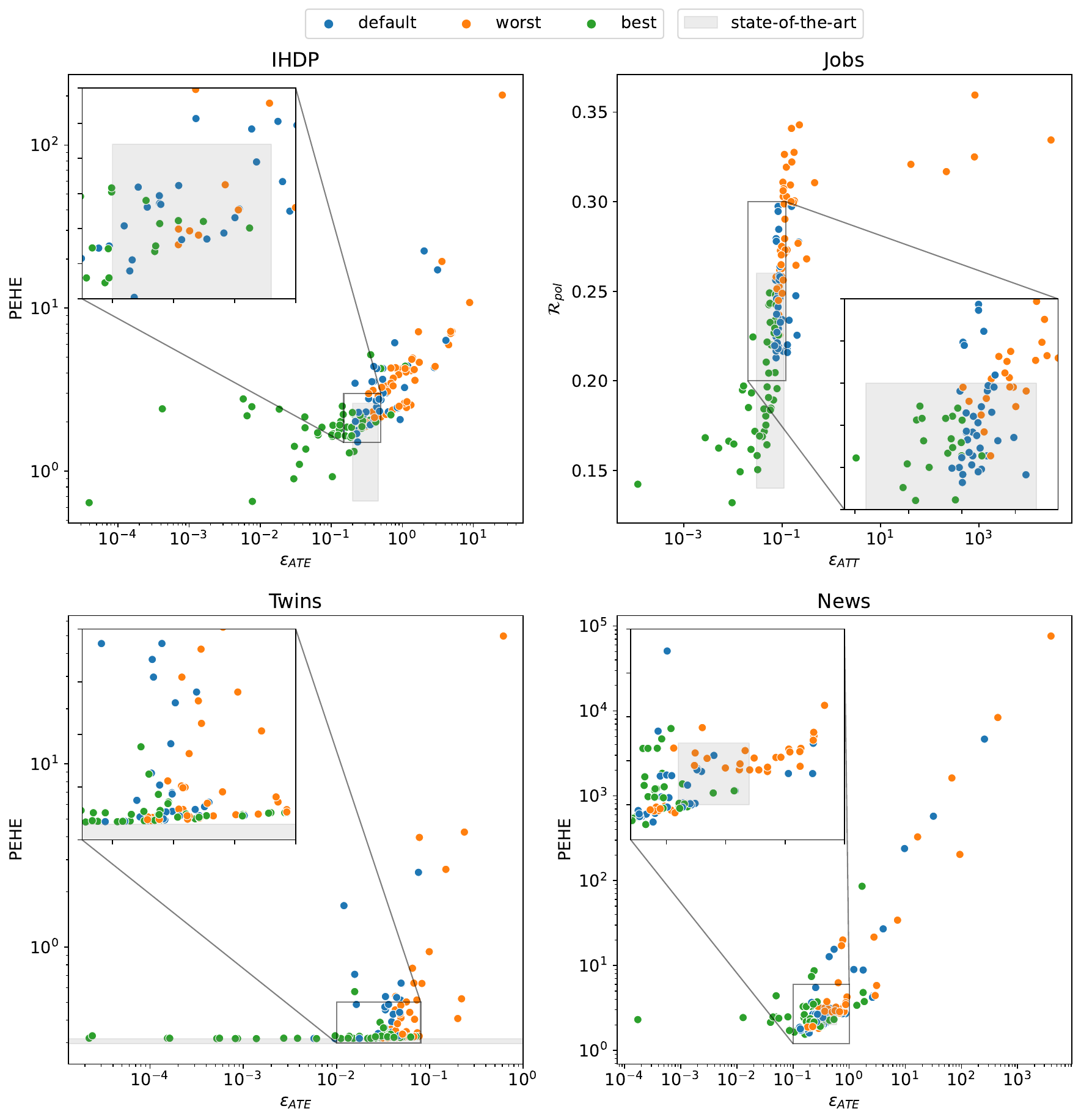}
    \caption{Performance of CATE estimators with three different types of quality of hyperparameters across all four datasets. Each data point represents the mean across dataset iterations. Lower is better applies to all metrics, thus the closer to bottom-left corner, the better. Note the logarithmic scale for all axes to aid visual presentation except policy risk in Jobs. Each of the potential metrics (see axis labels) was used \textbf{once} here: to select the best/worst hyperparameter values per estimator on the \textbf{test} data, and these performances are shown here.}
    \label{fig:c5}
\end{figure}

\begin{figure}[htb]
    \centering
    \includegraphics[width=0.90\textwidth]{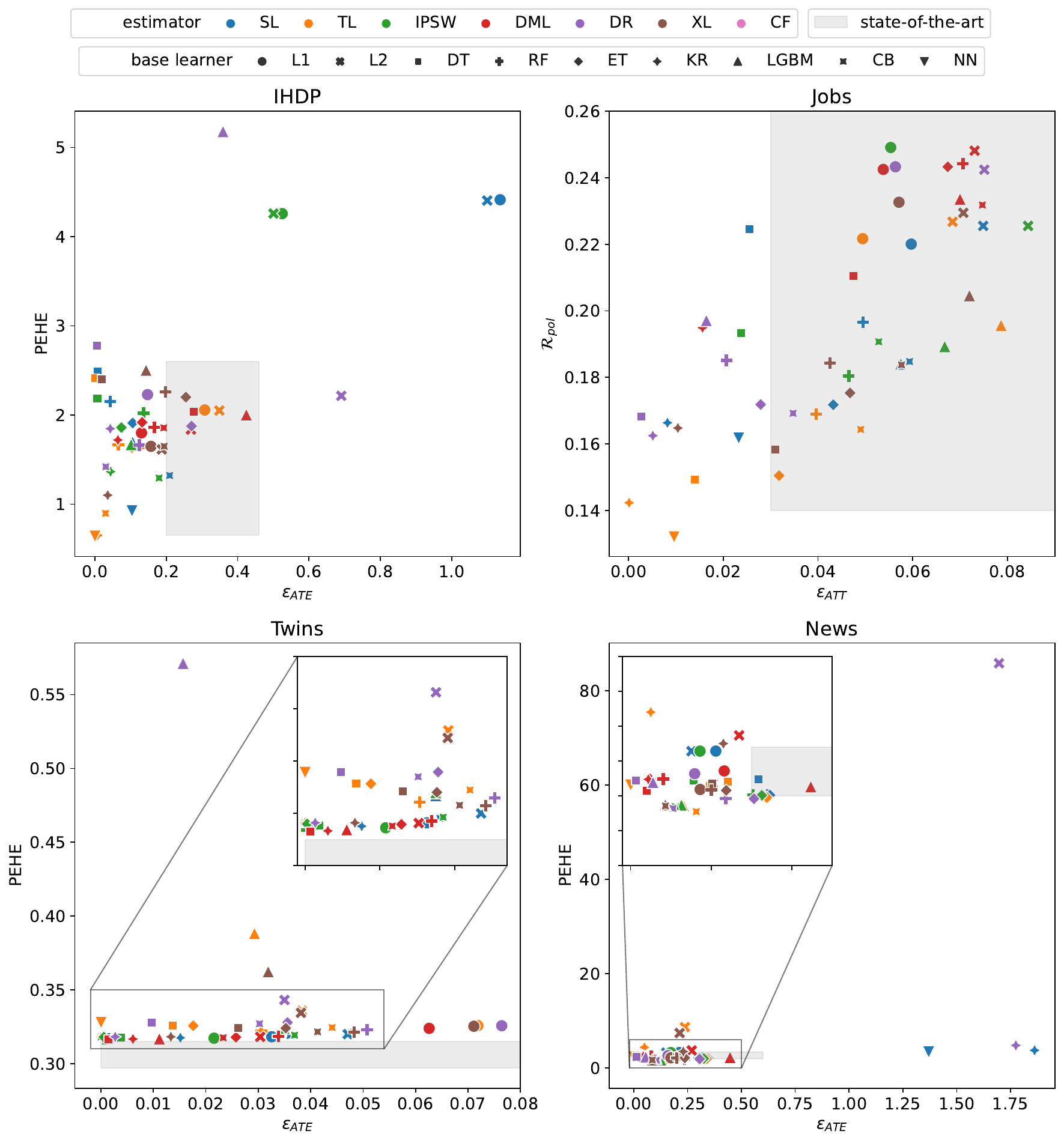}
    \caption{Oracle performances of CATE estimators across all four datasets, grouped by types of causal estimators and base learners. Each data point represents the mean across dataset iterations. Lower is better applies to all metrics, thus the closer to bottom-left corner, the better. Each of the potential metrics (see axis labels) was used \textbf{once} here: to select the best hyperparameter values per estimator on the \textbf{test} data, and these performances are shown here.}
    \label{fig:c6}
\end{figure}

\begin{figure}[htb]
    \centering
    \includegraphics[width=0.90\textwidth]{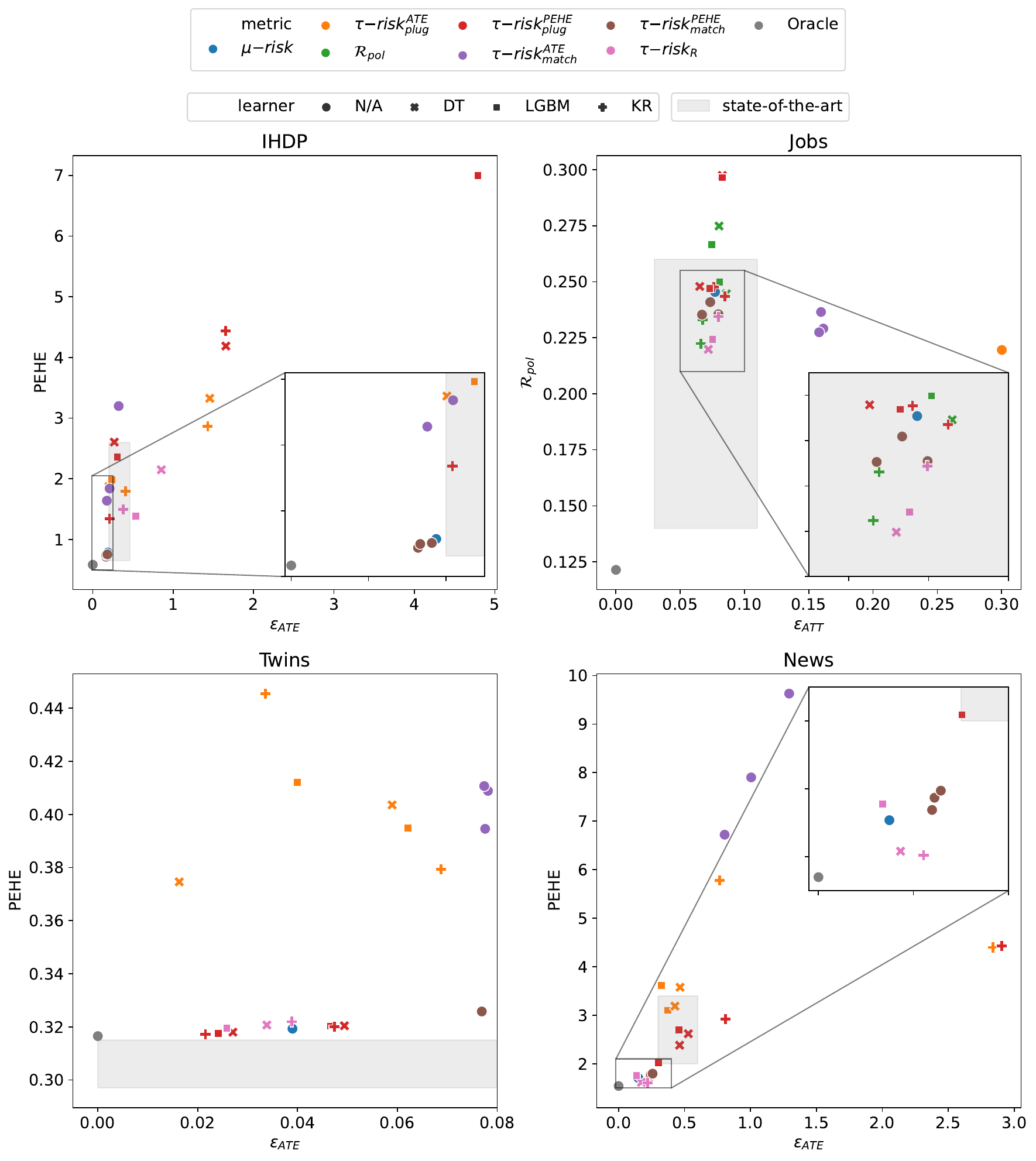}
    \caption{Performance of model selection metrics across all four datasets. Each data point represents the mean across dataset iterations. Lower is better applies to all metrics, thus the closer to bottom-left corner, the better. Each observable metric (see legend) was used to select the best candidate model on the validation data. Then, the potential metrics (see axis labels) were used to measure performances of selected candidates on the test set (shown here). The Oracle points represent the use of the potential metrics to select candidates on the \textbf{test} data. When using a potential metric to select candidates, the same metric is then used to measure the test performance. Note that some combinations of colours and shapes may appear more than once. This is the case for matching-based ($k\in\{1,3,5\}$) and plugin-based (S- and T-Learning) metrics due to multiple implementation variants explored here.}
    \label{fig:c7}
\end{figure}

Figures~\ref{fig:c5}-\ref{fig:c7} correspond to the results presented in Figures~\ref{fig:c2}-\ref{fig:c4} but use the alternative Oracle selection, as defined in Equations \eqref{eq:app1}, \eqref{eq:app2}, and \eqref{eq:app3}.


\end{document}